\documentclass[pdflatex,sn-nature]{sn-jnl}

\usepackage{graphicx}
\usepackage{multirow}
\usepackage{amsmath,amssymb,amsfonts}
\usepackage{booktabs}
\usepackage{array}
\usepackage{longtable}
\usepackage{xcolor}
\usepackage[breakable]{tcolorbox}
\usepackage{textcomp}
\usepackage{placeins}
\usepackage{capt-of}
\usepackage{bibunits}
\usepackage{eso-pic}
\bibpunct{}{}{,}{s}{}{\textsuperscript{,}}

\newcommand{\mainref}[1]{\ref{#1}}
\newcommand{\mainhyperref}[2]{\hyperref[#1]{#2}}
\newcommand{\suppref}[1]{\ref{#1}}
\newcommand{\supphyperref}[2]{\hyperref[#1]{#2}}

\begin{document}

  % \AddToShipoutPictureFG*{%
  %   \AtPageUpperLeft{%
  %     \put(\LenToUnit{\paperwidth},0){%
  %       \put(-\LenToUnit{50mm},-\LenToUnit{25mm}){%
  %         \makebox(0,0)[rt]{%
  %           \begin{tabular}{@{}r@{}}
  %             \small\sffamily Scientific Computing and Intelligence Group\\[-1pt]
  %             \small\sffamily Scaling Group
  %           \end{tabular}%
  %         }%
  %       }%
  %     }%
  %   }%
  % }
  
\articletype{Article}

\begin{bibunit}[sn-nature]
\title[A foundation model of numerical intelligence with cross-disciplinary generalization]{A foundation model of numerical intelligence with cross-disciplinary generalization}

\author[1]{\fnm{Chenghan} \sur{Wu}}

\author[1]{\fnm{Zongmin} \sur{Yu}}

\author*[1]{\fnm{Liu} \sur{Yang}}\email{yangliu@nus.edu.sg}

\affil[1]{\orgname{Department of Mathematics, National University of Singapore}, \orgaddress{\city{Singapore}, \country{Singapore}}}

\abstract{
Intelligence is commonly understood as the ability to acquire and apply knowledge, adapt to unfamiliar situations and solve new problems\cite{russell2021artificial,legg2007universal}. Large language models exhibit this capacity by inferring task-relevant knowledge from textual context and applying it to new tasks\cite{brown2020language}. Yet intelligence need not be confined to language. For scientific and social systems, we need models that acquire and apply knowledge from numerical context--an ability we call numerical intelligence. Here we introduce UNified In-Context Operator Networks (UNICON), a foundation model that exhibits numerical intelligence across disciplines. Using graph-based examples from a system as context, UNICON infers the predictive relation shared across them and applies it to queries from the same system. Across scientific and social systems, including those from disciplines absent from training, the same model approaches specialist performance without retraining. Combining UNICON with language-model agents to  perform contextual ensemble learning (CEL) yields further gains, enabling it to surpass state-of-the-art specialists in a discipline unseen during training. We further show that training-corpus diversity improves generalization to unseen disciplines. Together, these results establish UNICON as a foundation model of numerical intelligence and position it as a building block for a broader ecosystem of artificial intelligence.
}

\keywords{Foundation model, In-context learning, Graph neural network, Cross-disciplinary generalization}

\maketitle

\begin{quote}
\small
\itshape
There are, indeed, things that cannot be put into words. They make themselves manifest.

\normalfont\hfill Ludwig Wittgenstein, \emph{Tractatus Logico-Philosophicus}, 6.522\nocite{wittgenstein1961tractatus}~[\citenum{wittgenstein1961tractatus}]
\end{quote}

Scientific and social systems often make themselves known before they are put into words. Traffic flows across road networks, water moves through river basins, and atmospheric and geomagnetic states evolve over space and time. We encounter these systems through numerical observations organized as graphs, fields and multivariate sequences. These observations contain relations that may be learned before they are named, formalized or distilled into a specialist model.

Intelligence is commonly understood as the ability to acquire and apply knowledge, adapt to unfamiliar situations and solve new problems\cite{russell2021artificial,legg2007universal}. Large language models (LLM) make this capacity tangible: after broad textual training, they can acquire task-relevant knowledge and skills from a prompt and apply them to a new task\cite{devlin2019bert,brown2020language,hoffmann2022training}. Their success suggests a principle broader than language: context can provide a fixed model with the knowledge needed for a new problem and guide how that knowledge is applied. The question is whether numerical observations can serve the same role for scientific and social systems. We refer to the ability to acquire and apply knowledge from numerical context as \emph{numerical intelligence}.

Realizing numerical intelligence is difficult because numerical systems do not share an obvious vocabulary. A traffic graph and an atmospheric field may differ in their variables, units, sampling rates, geometry and prediction targets. Numerical models therefore tend to be designed for one system at a time; moving to a new system usually entails fitting a new model. Machine learning methods have been developed to approximate solutions, identify governing dynamics and learn solution operators for scientific systems\cite{brunton2016discovering,sirignano2018dgm,e2018deepritz,long2018pdenet,raissi2019physics,brunton2020machine,lu2021learning,e2020machine,li2021fourier,karniadakis2021physics,list2022learned,dehoop2022cost}. Graph-based models provide a natural representation for irregular domains and interacting systems\cite{bertozzi2012diffuse,bronstein2017geometric,kipf2018neural,battaglia2018relational,sanchezgonzalez2020learning,pfaff2021learning,brandstetter2022message}. Recent in-context operator learning takes a step towards this goal: given examples that pair input and output functions, a frozen model infers the relation they express and applies it to a new query\cite{yang2023context,yang2025fine,liu2023domain,yang2024pde,chen2024data,cole2024provable,serrano2025zebra,cao2026vicon,zhang2025probabilistic,wu2026graph,yang2026chop}. Existing demonstrations, however, have largely remained within individual operator families or closely related physical systems. Whether the same principle can unite cross-disciplinary numerical systems remains open.

Here we introduce UNified In-Context Operator Network (UNICON), a foundation model of numerical intelligence with cross-disciplinary generalization. 
UNICON is trained on data spanning hydrology, traffic, power systems, weather, land, ocean, soil, solar resources, and human mobility, with all systems presented in a shared graph-based context format. Within this format, the context consists of examples pairing observed histories with future states at a chosen forecast horizon, while the query is a new observed history whose state at the same horizon is to be predicted. At inference, UNICON adapts to a new system by using examples from that system as context, while its weights remain fixed.

\begin{figure}[!t]
\centering
\includegraphics[width=\linewidth]{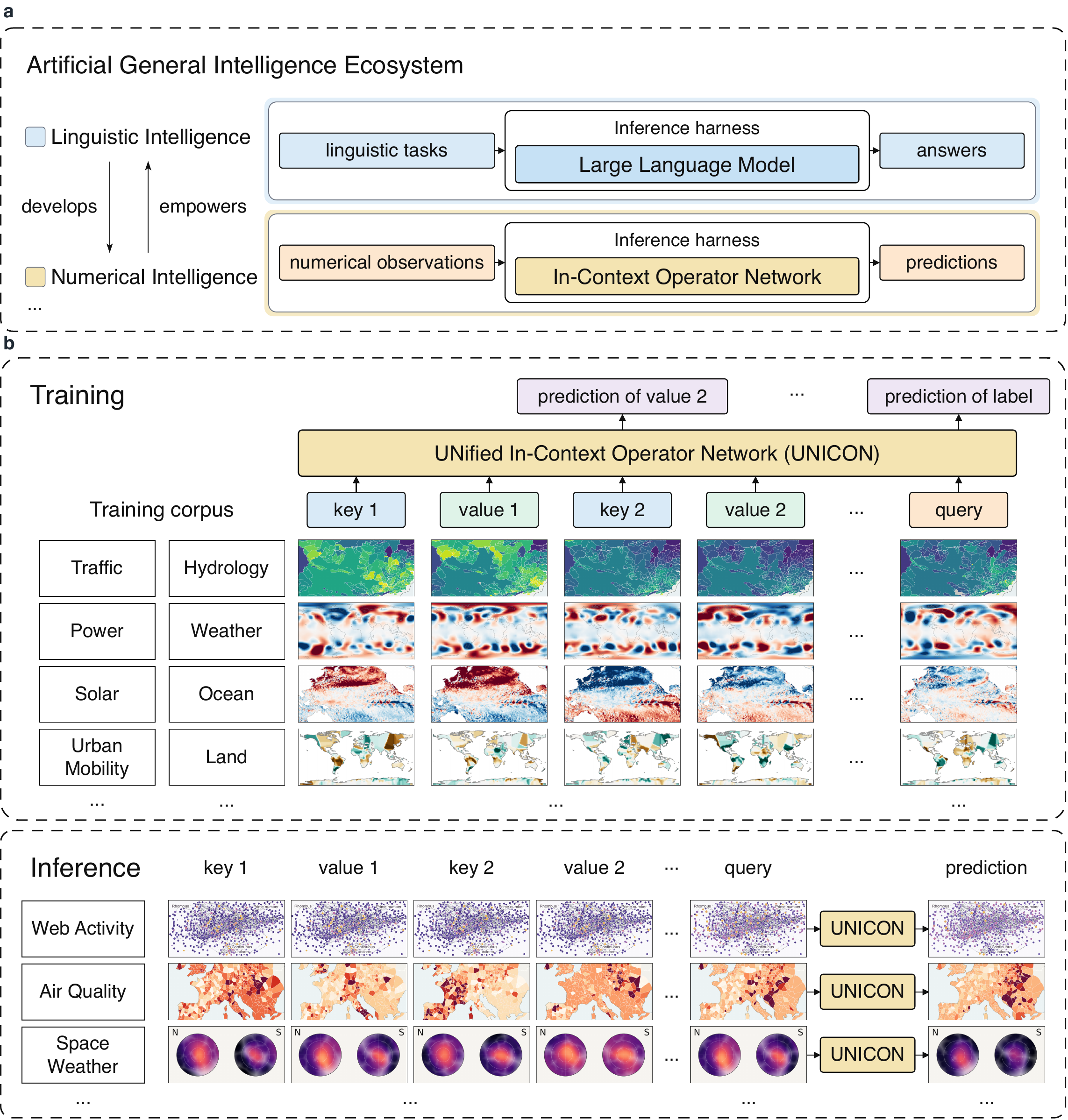}
\caption{\textbf{UNICON as a foundation model of numerical intelligence with cross-disciplinary generalization.} 
\textbf{a}, {Synergy of linguistic and numerical intelligence in an artificial-general-intelligence ecosystem.} 
Both intelligence paradigms are conceptualized as systems pairing a foundation model with an external inference harness, defined as an external program that orchestrates model calls without weight updates. Within this collaborative loop, linguistic intelligence strategically helps develop numerical intelligence, which in turn grounds and empowers linguistic intelligence with precise quantitative predictions.
\textbf{b}, {Training and Inference of UNICON.} 
UNICON is trained on a highly diverse, multi-source corpus spanning hydrology, traffic, power systems, and other disciplines, with all physical and social dynamics represented in a shared graph-based context format. Each training example pairs an observed history (key) with its future state (value) at a designated forecast horizon. At inference time, the frozen UNICON model adapts to entirely unseen disciplines (e.g., web activity, air quality, space weather) purely by learning from a few task-specific contextual examples, bypassing the need for parameter updates.
}
\label{fig:protocol}
\end{figure}

We evaluate the same frozen model on testing tasks spanning scientific and social systems, including three systems from disciplines entirely absent from training (air quality, web activity and space weather). Without any retraining, UNICON beats k-nearest neighborhood (kNN) baseline and approaches models trained specifically for each task. Remarkably, on web activity, UNICON inference reaches specialist-level performance even though neither the dataset nor the discipline was represented during training. Employing an inference-time harness of contextual ensemble learning (CEL) designed by LLM agent yields further gains, allowing UNICON to surpass state-of-the-art specialists in the web activity task.

To understand how contextual adaptation supports this performance, we systematically intervene on the supplied examples, showing that UNICON infers and uses the predictive relations expressed by contextual examples, rather than benefiting from the mere presence or length of context. Moreover, corpus-diversity experiments show that training across diverse numerical systems improves transfer to unseen systems and disciplines while retaining strong performance on training-corpus systems.

Together, these results establish UNICON as a foundation model for numerical intelligence: a single frozen model that infers system-specific predictive relations from numerical context and applies them to new queries across scientific and social systems. More broadly, they suggest that artificial general intelligence may emerge not from a single monolithic model, but from a collaborative ecosystem integrating linguistic and numerical intelligence, as is illustrated in Fig.~\ref{fig:protocol} a.

\enlargethispage{2\baselineskip}

\section*{Unifying disciplines with graph contexts}

Generalizing a single model across disciplines poses two challenges. First, the
model must accommodate systems and tasks that differ in numerical scale, numbers and sets of variables, geometries, observation-history length, and forecast horizon. Second, it must adapt to how the entities and variables of a system interact and how observed histories map to the requested future state. UNICON addresses these challenges with a specially designed network architecture and in-context learning paradigm.

In UNICON, each system is represented as a graph whose nodes denote system-specific entities, such as river catchments, geomagnetic grid cells or measurement stations, and webpages. Each node carries one or more system-specific variables observed over time, for example, meteorological forcings and streamflow for a catchment, multiple field components at a geomagnetic site, or page-view activity for a webpage. The graph encodes relations among entities, whereas the variable axis represents the quantities associated with each entity.

The graph provides a common representation of relations among entities, but it does not fully specify the dynamics of a system or the prediction task.   Constructing a cross-disciplinary vocabulary for explicitly describing this information would be impractical; more fundamentally, the underlying dynamics are often only partially understood and cannot themselves be fully specified. UNICON therefore supplies this information through numerical examples, allowing the frozen model to adapt in context without system or task identifiers or parameter updates.

In particular, UNICON receives a sequence $[K_1,V_1,\ldots,K_D,V_D,K_q]$ (Fig.~\ref{fig:protocol} b), whose elements are defined over the same graph. Each key $K_i$ is an observed history, and each value $V_i$ is its corresponding future state at a chosen forecast horizon. These key-value pairs reveal how entities and variables interact within each state and how observed histories relate to future states. UNICON processes the sequence in an autoregressive way: it predicts the corresponding value from the current key and the preceding key-value pairs, and at the final query it predicts the withheld future state $V_q$. 

UNICON combines a common graph sequence interface with an attention-based architecture. The same architecture operates across datasets and flexible contextual example cardinalities, without dataset-specific input or output layers. Architectural details are provided in Supplementary Material Section~\suppref{sec:unicon-architecture}.

\section*{Cross-disciplinary generalization}

UNICON contains 41.6M trainable parameters and was trained once on
20 data sources spanning hydrology, traffic, power systems, weather, land, ocean, soil, solar resources and human mobility. Together, these sources comprise approximately 216GB of numerical data. During training, contextual examples were sampled randomly from a precomputed pool of candidates similar to the query (Supplementary Material Section~\suppref{sec:supp-context-retrieval}). Training used a total compute budget of $9.6\times10^{17}$ floating-point operations and ran for 28 hours on 2 NVIDIA H200. The resulting model was frozen and used unchanged throughout evaluation. 

We evaluated this single model on nine systems at three levels of separation from the training corpus: two unseen periods from training sources, four unseen datasets from represented disciplines and three systems from disciplines absent from training. The last group comprised air quality, web activity and space weather. Any task-specific adaptation occurred through contextual examples.

\begin{figure}[t!]
\centering
\makebox[\linewidth][c]{%
\hspace*{0.025\linewidth}%
\includegraphics[width=\linewidth]{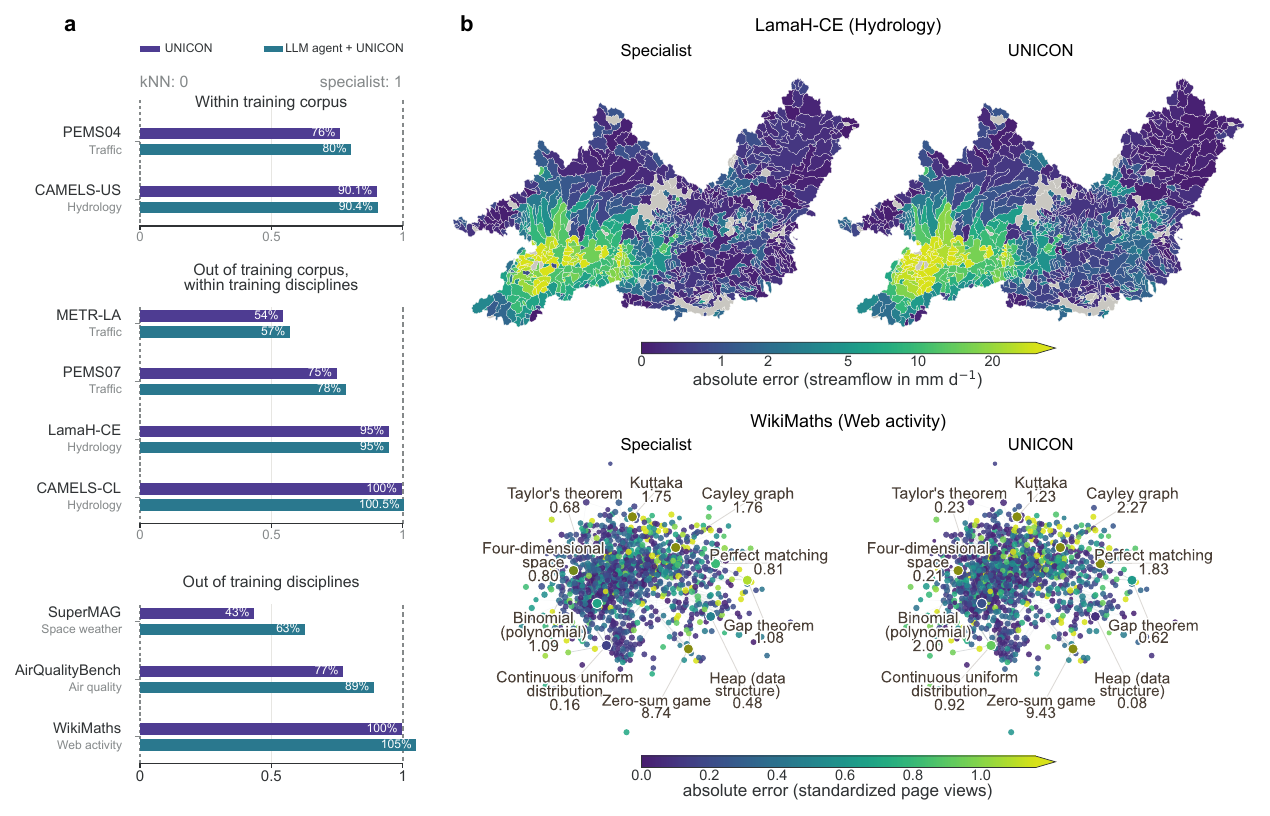}%
\hspace*{-0.025\linewidth}}
\begin{minipage}{0.95\linewidth}
\caption{\textbf{One frozen model generalizes across disciplines, with further gains from LLM-agent assistance.}
\textbf{a,} Performance of UNICON direct inference (purple) and contextual ensemble learning orchestrated by LLM-agents (blue), both using the same frozen model, across evaluation systems spanning unseen periods from training sources, unseen datasets from represented disciplines and systems from disciplines absent from training. The evaluation metrics are normalized to the range between 0 (kNN baseline) and 1 (task-specific specialist).
\textbf{b,} Absolute error of one-day-ahead prediction for LamaH-CE (top) and WikiMaths (bottom), on a representative test case. Top, streamflow error across LamaH-CE river catchments in Central Europe\cite{klingler2021lamah} for an LSTM specialist (left) and UNICON (right). Bottom, error in standardized page views across the WikiMaths graph of mathematics-related Wikipedia pages for a LAMP specialist\cite{castelli2026adaptive} (left) and UNICON (right).}
\label{fig:prediction-views}
\end{minipage}
\end{figure}

For each of the nine systems, we compared UNICON with a kNN baseline and a
leading task-specific specialist~\cite{liu2023staeformer,shao2022d2stgnn,
zhang2024eddlstm,castelli2026adaptive,wu2019graphwavenet}. The kNN baseline used the same five retrieved contextual examples as UNICON, with weights calculated by cosine similarity. Figure~\ref{fig:prediction-views}a shows UNICON's performance relative to the kNN baseline and task-specific specialist on each system.

We first evaluate direct UNICON inference with a straight forward context construction, where the examples are retrieved by selecting ones with keys of the highest similarity to query.

Across all evaluation tasks, UNICON improves
substantially over the kNN baseline (Fig.~\ref{fig:prediction-views}a). Most notably, on WikiMaths, direct UNICON reaches specialist-level performance even though neither the dataset nor the broader discipline of web activity was represented during training. 

To look beyond the aggregate scores, we examine LamaH-CE and WikiMaths in
greater detail. These systems represent two levels of generalization:
LamaH-CE is a new dataset from a discipline represented during training,
whereas WikiMaths belongs to a discipline absent from training. We examine
when and where prediction errors occur and how performance changes with
forecast horizon.

For LamaH-CE, the aggregate score masks differences in relative accuracy
across flow conditions and forecast horizons. The specialist is more accurate
during high-flow events and at the shortest forecast horizon, whereas UNICON
performs better at lower flows and intermediate horizons. Neither method
consistently dominates at the longest horizon
(\supphyperref{fig:lamah-flow-regimes}{Extended Data
Fig.~\suppref{fig:lamah-flow-regimes}}). These differences indicate
complementary strengths rather than a consistent ranking between the two
models.

\enlargethispage{2\baselineskip}
WikiMaths provides the stronger test of cross-disciplinary generalization.
Despite having encountered neither the dataset nor its discipline during
training, the frozen UNICON model performs within the range of a
state-of-the-art specialist trained specifically on WikiMaths. Their errors are
of similar scale across both aligned test dates and high-activity regions of
the graph
(Fig.~\ref{fig:prediction-views}b;
\supphyperref{fig:wikimaths-uncertainty}{Extended Data
Fig.~\suppref{fig:wikimaths-uncertainty}}). This comparison also persists
beyond the next-day setting: the methods remain comparable across shorter
horizons, while UNICON becomes more accurate at the longest evaluated horizons
(\supphyperref{fig:wikimaths-leadtime}{Extended Data
Fig.~\suppref{fig:wikimaths-leadtime}})\cite{castelli2026adaptive}.

Together, these case studies show that UNICON's competitive performance is not confined to a small set of favorable cases or a single
evaluation setting. It persists across temporal,
spatial and horizon-specific analyses.

\section*{LLM-orchestrated contextual ensemble learning}

A foundational insight in in-context operator learning is that the same physical problem can be addressed in different ways by reformulating its numerical prompt. As first demonstrated by the change-of-variables and varying-stride techniques~\cite{yang2024pde}, a frozen in-context operator network can yield distinct prediction behaviors when evaluated on different equivalent representations of the same underlying system~\cite{yang2024pde}. This discrepancy opens up a novel paradigm for optimizing inference purely in the prompt space via an inference-time ``harness'', defined as an external program wrapping around the frozen model to systematically orchestrate model calls without weight updates~\cite{yang2026chop}. The Chain-of-Operators (CHOP) framework~\cite{yang2026chop} represents an instantiation of this paradigm, implementing a harness family that routes numerical prompts specifically through explicit, closed-form elementary operator transformations~\cite{yang2026chop}.

The concept of an inference-time harness, however, extends far beyond CHOP's algebraic reformulations applied to a static context. In this work, we introduce another fundamental capability of an inference harness: actively retrieving and curating contextual examples from a vast database of available observations. Because complex systems exhibit multifaceted dynamics, there is no single optimal way to construct a context; rather, different retrieval strategies, such as seasonal matching or scale alignment, capture complementary physical or structural aspects of the query.

Orchestrating these diverse context-building pathways and fusing their predictions effectively operationalizes the harness as an ensemble-learning system, a paradigm we term ``contextual ensemble learning'' (CEL). While traditional ensemble learning relies on the costly training of multiple independent networks, CEL elicits collective predictive robustness from a single, completely frozen foundation model purely by diversifying its context. Moreover, unlike prompt ensembling in large language models that aggregates semantically redundant reformulations of a static task, CEL addresses the multifaceted nature of complex systems. By constructing contexts from distinct observation slices, CEL systematically exposes the frozen network to complementary facets of the system, fusing these partial views to approximate the target operator far more reliably.

Crucially, such active context curation cannot be solved by rigid numerical algorithms alone, as it requires interpreting task specifications in texts and domain-specific heuristics. This requirement marks the natural entry point for Large Language Models (LLMs), enabling a powerful synergy between linguistic intelligence, which excels at semantic reasoning and strategic orchestration, and numerical intelligence, as realized in the frozen UNICON model.

Specifically, for the WikiMaths task, the LLM-orchestrated CEL first queries the frozen in-context operator network to generate a baseline prediction using the default top-similarity retrieval rule. It then make auxiliary predictions using different groups of contextual examples, including examples from the same day of the week, those matched to the scale of the query’s recent observations, and a broader set of similarity-retrieved examples. Rather than simply averaging the auxiliary predictions, the designed CEL treats their disagreement as an indicator of uncertainty and uses it to decide how much to adjust the baseline prediction dynamically for each query, yielding the final forecast (Supplementary Material Section~\suppref{sec:supp-agent-assisted-inference}).

Direct comparison reveals that CEL consistently enhances UNICON's capabilities across most evaluation tasks (Supplementary Material Section~\suppref{sec:supp-evaluation}). Remarkably, UNICON with CEL even exceeded the performance of task-specific specialists on CAMELS-CL and WikiMaths, outperforming the state-of-the-art specialist trained exclusively for the latter (Fig.~\ref{fig:prediction-views}a). 

These gains, achieved entirely without parameter updates for UNICON, highlight a defining flexibility of in-context operator networks: varying how context is constructed can adapt model behavior at inference, demonstrating a clear path toward test-time compute scaling through repeated model calling. This paradigm also instantiates a powerful synergy between linguistic and numerical intelligence. Within this collaborative framework, language agents interpret high-level task specifications to strategically coordinate the inference pipeline and design the adaptive harness for the in-context operator network, while the in-context operator network, acting through this harness, empowers linguistic intelligence to solve complex scientific and physical problems far beyond the reach of language alone.

\FloatBarrier

\noindent\makebox[\linewidth][c]{%
\includegraphics[width=\linewidth]{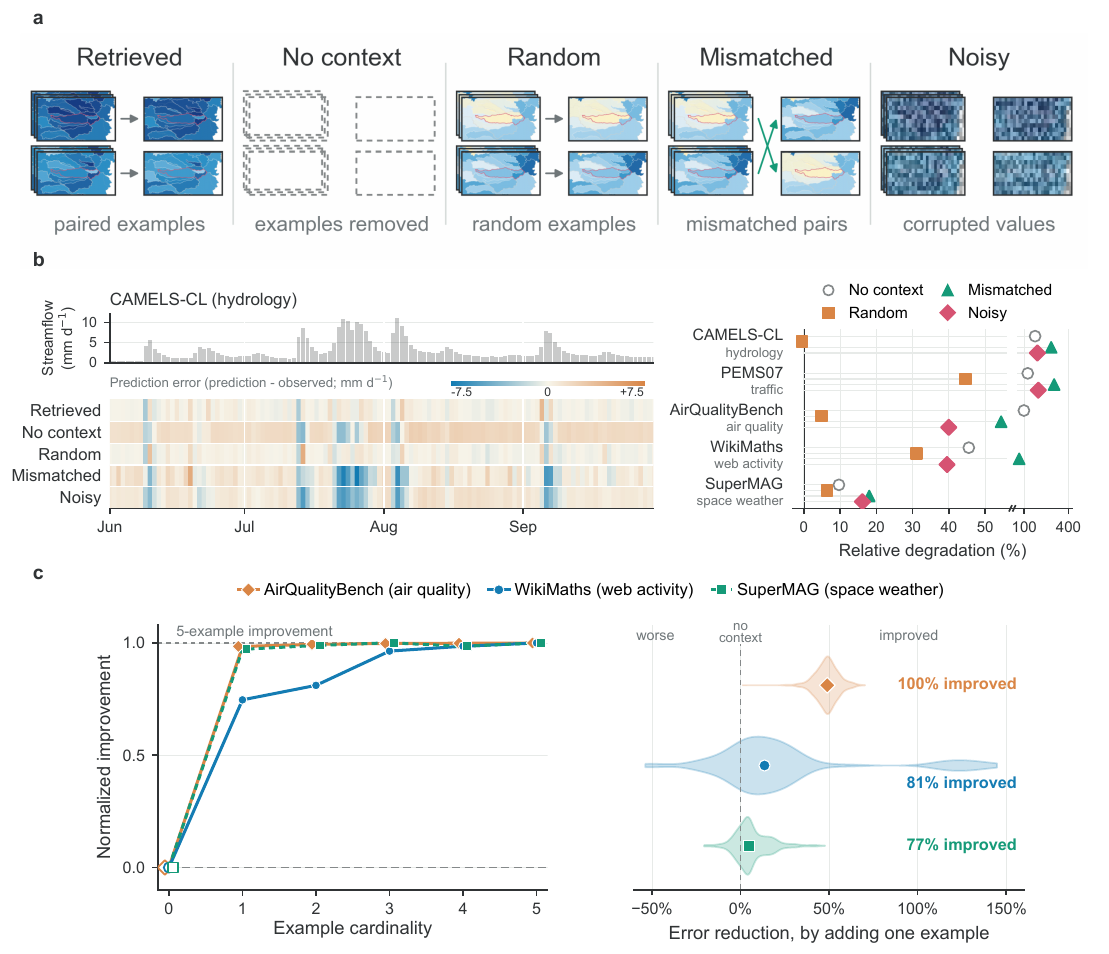}}
\captionof{figure}{\textbf{Numerical examples support cross-disciplinary generalization.}
\textbf{a,} Five inference conditions: examples retrieved by similarity; no
examples; randomly sampled but correctly paired examples; mismatched
history--future pairs; and examples with corrupted numerical values.
\textbf{b,} Observed streamflow and prediction error under each condition
during a representative unseen CAMELS-CL event (left), and percentage change
in each dataset's task-specific error relative to five retrieved examples
across five datasets (right). Positive values indicate worse performance.
\textbf{c,} Left, improvement as the number of retrieved examples increases,
normalized to the range between 0 (no context) and 1 (five retrieved examples).
Right, distributions across queries of the error reduction obtained from one
retrieved example, normalized by each dataset's aggregate no-context error;
markers denote medians and labels give the percentage of queries improved.}
\label{fig:example-controls}

\section*{Learning from contextual examples}

UNICON generalizes across disciplines when supplied with numerical examples at
inference. Here we tested whether this adaptation comes from learning from the
examples themselves, rather than merely a response to the presence or length of
context. Holding the frozen model and query fixed, we systematically changed only the contextual examples. In particular, we compared examples retrieved by similarity with no context, randomly sampled but correctly paired examples, mismatched history-future pairs and examples
whose numerical values were corrupted (Fig.~\ref{fig:example-controls}a).

The supplied examples consistently affected prediction. Removing them,
breaking the correspondence between observed histories and subsequent states,
or corrupting their numerical values increased error across all five target
systems. Randomly sampled but correctly paired examples outperformed these
three controls, whereas examples retrieved by similarity (Supplementary Material Section~\suppref{sec:supp-context-retrieval}) performed best overall
(Fig.~\ref{fig:example-controls}b). In the representative hydrology event,
retrieved examples also produced forecasts that more closely followed the
observed rises and peaks. These results show that UNICON learns from the
numerical relations expressed by the examples and benefits further when those
examples are relevant to the query.

Strikingly, much of the cross-disciplinary gain appears with only one example.
Across all three disciplines absent from training, a single retrieved example
recovers most of the improvement obtained with five and lowers error for most
queries (Fig.~\ref{fig:example-controls}c). Although counted as a single example, each pairs a graph-wide, multivariate observation sequence with a corresponding graph-wide future state, providing a large collection of jointly structured inputs and targets across nodes and variables. These results show that UNICON can achieve substantial generalization gains in disciplines absent from training using only one or a few relevant numerical examples.

\begin{figure}[p]
\centering
\makebox[\linewidth][c]{%
\includegraphics[width=\linewidth]{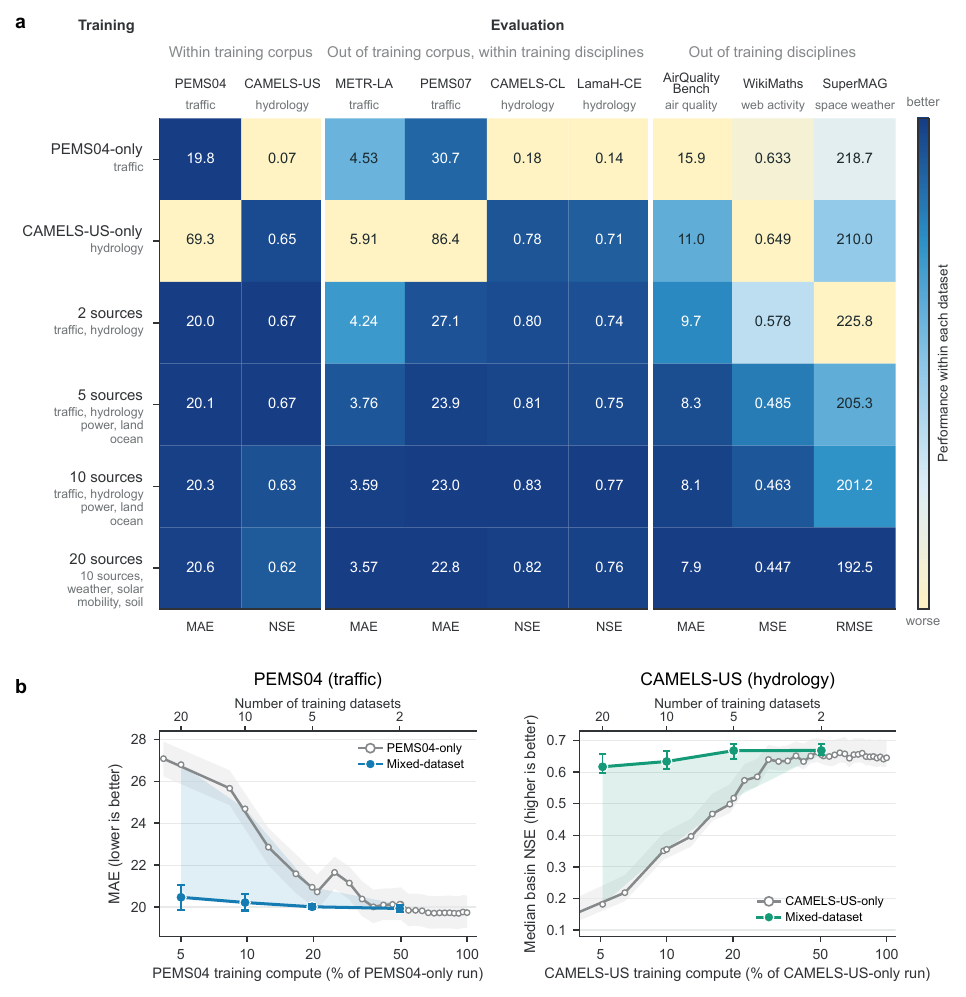}}
\caption{\textbf{Training-corpus diversity improves cross-disciplinary generalization while preserving performance on its training sources.}
\textbf{a,} Performance of models with identical architecture and size trained under the same total budget on corpora containing 1, 2, 5, 10 or 20 data sources. Columns group evaluation systems into those represented in the training corpus, those outside the training corpus but from represented disciplines, and those from unseen disciplines. Colours are normalized separately within each evaluation dataset; numbers report the corresponding metric indicated below each column. \textbf{b,} PEMS04 MAE (left) and CAMELS-US median basin NSE (right) for the learning trajectories of models trained on a single source (grey) and models trained on more diverse corpora (coloured). The lower axes express training on PEMS04 or CAMELS-US as a percentage of a complete run using only that source; the upper axes show the corresponding number of sources in the training corpus. Coloured shading marks the performance difference between each model trained on a more diverse corpus and the corresponding single-source trajectory after comparable training on that source. Grey bands and coloured error bars indicate 95\% bootstrap intervals.}
\label{fig:source-diversity}
\end{figure}

\section*{Generalization improvements through corpus diversity}

The broad capabilities of large language models are built on pretraining
  corpora that span many textual domains. This raises an analogous question for
  numerical foundation models: whether exposure to a wider range of numerical
  systems improves their ability to learn from context in unseen systems and
  disciplines. We therefore trained models with the same architecture, parameter count and total compute on corpora containing between one and 20 data sources
  (Fig.~\ref{fig:source-diversity}a).
  
Models trained on a single source specialized strongly in their training system but generalized poorly elsewhere. As the training corpus became more diverse, performance improved on systems outside the training corpus, including those from unseen disciplines, while largely preserving performance on systems represented during training. The model trained on the most diverse corpus consequently achieved the strongest overall performance across the evaluation set (Fig.~\ref{fig:source-diversity}a).

  Because broader corpora allocate less of the fixed training budget to each source, we compared mixed-corpus models with PEMS04-only and CAMELS-US-only models (Fig.~\ref{fig:source-diversity}b). In the 20-source model, only about one twentieth of the total budget was allocated to each
  source. A single-source model trained with the same one-twentieth budget performed substantially worse on both datasets, whereas the 20-source model remained close to the single-source model trained with the full budget. In other words, corpus diversity extends a single contextual model's reach across disciplines with little loss of performance on its training sources. This result makes adding further data sources a promising path towards more capable numerical foundation models.

\section*{Discussion}

The central finding of this work is that a fixed numerical intelligence model can adapt across disciplines by learning from context. In UNICON, training equips the model to learn from graph-based numerical examples, whereas context conveys the system- and task-specific knowledge needed for each query. The model infers the predictive relation shared across the examples and applies it to queries from the same system. By diversifying these contexts and fusing their predictions, this process establishes a paradigm of ``contextual ensemble learning''.

Without retraining, UNICON with contextual ensemble learning beats state-of-the-art specialist performance even on systems from disciplines absent from training and operates across systems that differ in their variables, units, geometry and prediction targets. This ability to acquire and apply knowledge from numerical context is the basis of UNICON's numerical intelligence.

Contextual interventions clarify how this adaptation occurs. Performance is higher with randomly sampled but correctly paired examples than with no context or with contexts containing mismatched key-value pairs or corrupted numerical values; retrieving examples for their relevance to the query provides an additional improvement. Together, these controls show that UNICON infers and uses the predictive relations expressed by contextual examples, rather than benefiting from the mere presence or length of context.

Under a fixed training budget, training on a broader range of numerical systems improves performance on unseen systems and disciplines while largely maintaining performance on systems included in training. This pattern indicates that corpus diversity strengthens UNICON's ability to learn from contextual examples in new systems.

The present study has several limitations. It evaluates numerical intelligence through forecasting; whether the same approach extends to imputation, anomaly detection, simulation or control remains to be tested. The corpus-diversity experiments compare particular corpus compositions under a fixed training budget and therefore do not establish a general scaling law relating performance to data volume, model size and compute. Finally, systematic methods for selecting and organizing contextual examples for different systems and tasks require further study.

More broadly, the integration of agent-assisted inference and frozen in-context operator networks provides a powerful template for the future of scientific artificial general intelligence. By pairing language-model agents with UNICON, we demonstrate a profound synergy between linguistic and numerical intelligence. In this collaborative framework, linguistic intelligence leverages semantic reasoning to interpret high-level task specifications, understand domain-specific metadata, and strategically coordinate the inference pipeline, while numerical intelligence, acting through adaptive harnesses, empowers linguistic intelligence to solve complex scientific and physical problems far beyond the reach of language alone. This division of labor suggests that AGI may emerge not from a single, monolithic, all-encompassing architecture, but rather from a collaborative, decentralized ecosystem of specialized linguistic and numerical building blocks, with UNICON providing a highly generalizable, retrain-free numerical foundation block.

\clearpage
\backmatter
\section*{Data availability}

All datasets used in this study are available from their original providers, subject to the providers' terms. The hydrological sources and targets were obtained from the Caravan v1.5 NetCDF archive (\href{https://doi.org/10.5281/zenodo.14673536}{Zenodo record 14673536}), which contains the CAMELS-US, CAMELS-BR, HYSETS, CAMELS-GB, CAMELS-CL and LamaH-CE subsets used here. Traffic data were obtained from \href{https://dot.ca.gov/programs/traffic-operations/mpr/pems-source}{Caltrans PeMS} and the public benchmark releases cited in the Methods. EIA-930 data are available through the \href{https://doi.org/10.5281/zenodo.16262491}{PUDL archive}; the European power-system source combines PyPSA-Eur topology with \href{https://doi.org/10.25832/time_series/2020-10-06}{Open Power System Data}. GLDAS is available from \href{https://doi.org/10.5067/E7TYRXPJKWOQ}{NASA GES DISC}; ERA5-Land and ERA5 are available through the \href{https://cds.climate.copernicus.eu/}{Copernicus Climate Data Store}, with the WeatherBench2 copy described in its \href{https://weatherbench2.readthedocs.io/en/latest/data-guide.html}{data guide}; and SMAP-L4 is available from the \href{https://doi.org/10.5067/T5RUATAQREF8}{National Snow and Ice Data Center}. The GLORYS and Mediterranean Sea products are available from the \href{https://data.marine.copernicus.eu/product/GLOBAL_MULTIYEAR_PHY_001_030/description}{Copernicus global-ocean} and \href{https://data.marine.copernicus.eu/product/MEDSEA_MULTIYEAR_PHY_006_004/description}{Mediterranean Sea} records. Solar and mobility data are available from \href{https://power.larc.nasa.gov/docs/services/api/temporal/hourly/}{NASA POWER}, \href{https://data.ny.gov/Transportation/MTA-Subway-Hourly-Ridership-2020-2024/wujg-7c2s}{New York State Open Data} and \href{https://citibikenyc.com/system-data}{Citi Bike System Data}.

The AirQualityBench files used here are available from \href{https://huggingface.co/datasets/xuxing123/aq_dataset}{Hugging Face}; WikiMaths is distributed by PyTorch Geometric Temporal as \href{https://raw.githubusercontent.com/benedekrozemberczki/pytorch_geometric_temporal/master/dataset/wikivital_mathematics.json}{the WikiMaths JSON record}; and the event definitions, processed Solar-Magnetic coordinate files and Gannon inputs used for the SuperMAG evaluation were obtained from the open GeoDGP-associated \href{https://doi.org/10.7302/6brp-0y03}{Deep Blue Data release}\cite{chen2024geodgpdata}. The underlying magnetic measurements originate from \href{https://supermag.jhuapl.edu/}{the SuperMAG service}.

Owing to data volume and provider licences, derived copies of the raw datasets are not redistributed.

\section*{Acknowledgements}

Liu Yang acknowledges support from the National Research Foundation,
Singapore, under the NRF fellowship (Project No. NRF-NRFF17-2025-0006). We
acknowledge NUS IT's Research Computing group for providing computational
support.

We gratefully acknowledge the teams and organizations that developed, curated
and made available the datasets used in this study: Caravan v1.5, including
CAMELS-US, CAMELS-BR, HYSETS, CAMELS-GB, CAMELS-CL and LamaH-CE; Caltrans PeMS
and the public releases of PEMS04, PEMS07, PEMS08 and METR-LA; EIA-930 and
PUDL; PyPSA-Eur and Open Power System Data; GLDAS, ERA5-Land, WeatherBench2
ERA5 and SMAP-L4; GLORYS for the North Pacific, Southern Ocean and Kuroshio,
and the Copernicus Mediterranean Sea reanalysis; NASA POWER data for the
United States and Australia; MTA Subway and Citi Bike data; AirQualityBench
and OpenAQ; WikiMaths and PyTorch Geometric Temporal; and the GeoDGP-associated
Deep Blue Data release. We also gratefully acknowledge the SuperMAG
collaborators
(\url{https://supermag.jhuapl.edu/info/?page=acknowledgement}).

% Journal-only section; intentionally omitted from the arXiv manuscript.
% \section*{Author contributions}

% Journal-only section; intentionally omitted from the arXiv manuscript.
% \section*{Competing interests}

% References remain outside the Discussion-onward scientific-content boundary.
\renewcommand{\refname}{References}
% Pre-generated from references.bib for the combined arXiv build.

\end{bibunit}

\clearpage
\phantomsection
\section*{Supplementary Material}
% Shared presentation adapter for the combined arXiv and standalone SI builds.
% The scientific Supplement source remains channel-independent.

% Keep citation links local to the Supplement when a key also appears in the
% main bibliography of the combined arXiv PDF.
\makeatletter
\def\@extra@binfo{.supp}
\def\@extra@b@citeb{.supp}
\makeatother

\setcounter{section}{0}
\renewcommand{\thesection}{\Alph{section}}
\renewcommand{\thesubsection}{\thesection.\arabic{subsection}}
\renewcommand{\thesubsubsection}{\thesubsection.\arabic{subsubsection}}
\renewcommand{\theHsection}{supp.\Alph{section}}
\renewcommand{\theHsubsection}{\theHsection.\arabic{subsection}}
\renewcommand{\theHsubsubsection}{\theHsubsection.\arabic{subsubsection}}

\setcounter{figure}{0}
\renewcommand{\thefigure}{\arabic{figure}}
\renewcommand{\theHfigure}{supp.\arabic{figure}}
\renewcommand{\figurename}{Extended Data Fig.}

\setcounter{table}{0}
\renewcommand{\thetable}{\arabic{table}}
\renewcommand{\theHtable}{supp.\arabic{table}}
\renewcommand{\tablename}{Supplementary Table}

\setcounter{equation}{0}
\renewcommand{\theequation}{S\arabic{equation}}
\renewcommand{\theHequation}{supp.\arabic{equation}}

% Reader-facing excerpts from the instructions supplied to the LLM agents.
% The restrained, breakable treatment preserves the source hierarchy without
% turning long Supplementary Methods material into numbered manuscript parts.
\newtcolorbox{agentinstructionbox}[1]{
  breakable,
  sharp corners,
  colback=white,
  colframe=black!55,
  colbacktitle=black!5,
  coltitle=black,
  boxrule=0.45pt,
  left=7pt,
  right=7pt,
  top=5pt,
  bottom=6pt,
  before skip=9pt,
  after skip=11pt,
  fonttitle=\small\bfseries,
  fontupper=\small,
  toptitle=3pt,
  bottomtitle=3pt,
  title={#1},
  title after break={#1\space\textnormal{(continued)}}
}

\newenvironment{agentprocedurebox}[1]
  {\begin{agentinstructionbox}{#1}}
  {\end{agentinstructionbox}}

% Compact journal-style pseudocode used for a reader-facing prediction
% procedure. The serif typography and restrained line labels distinguish the
% scientific algorithm from a raw source-code listing.
\newenvironment{agentproceduresteps}{%
  \par\noindent
  \begingroup
  \setlength{\tabcolsep}{0pt}%
  \renewcommand{\arraystretch}{1.18}%
  \begin{tabular}{@{}>{\raggedleft\arraybackslash}p{0.10\linewidth}@{\hspace{8pt}}>{\raggedright\arraybackslash}p{0.85\linewidth}@{}}%
}{%
  \end{tabular}%
  \endgroup
  \par
}

\newcommand{\agentprocedurestep}[1]{%
  \textcolor{black!55}{\textbf{#1}}%
}

\newcommand{\agentexcerptheading}[1]{%
  \par\noindent\textbf{#1}\par\nobreak\smallskip\nobreak
}

\newcommand{\agentexcerptlabel}[1]{%
  \par\noindent\textbf{#1}\par\nobreak\smallskip\nobreak
}

\newcommand{\agentexcerptdivider}{%
  \par\medskip\noindent\textcolor{black!25}{\rule{\linewidth}{0.35pt}}%
  \par\medskip
}

\setcounter{tocdepth}{2}
\renewcommand{\contentsname}{Contents}

\tableofcontents
\clearpage
\begin{bibunit}[sn-nature]
% Canonical reviewed Supplementary Material manifest.
% Publication back matter is owned by the main manuscript wrappers.
\section{Problem formulation}

UNICON is evaluated as a frozen model that infers a system-specific predictive
relation from numerical examples. For a system \(s\), let
\(x_t^{(s)}\in\mathbb{R}^{N_s\times C_s}\) denote its observed numerical state
at time \(t\). Here, \(N_s\) is the number of nodes in the system graph and
\(C_s\) is the number of channels, each corresponding to one system-specific
variable. Depending on the system, a node may
correspond to a grid cell, measurement station, region, catchment or another
system-specific entity. Each task specifies a history length \(T_s\), a
forecast horizon \(\Delta_s\) and a set of examples.

Each example is a key-value pair. The key
\(K_i=x_{t_i-T_s+1:t_i}^{(s)}\in\mathbb{R}^{T_s\times N_s\times C_s}\) is an
observed numerical history, and the value
\(V_i=x_{t_i+\Delta_s}^{(s)}\in\mathbb{R}^{N_s\times C_s}\) is the corresponding
future state. Given \(D\) examples
\(\mathcal{C}_q=\{(K_i,V_i)\}_{i=1}^{D}\) and a query
\(K_q=x_{t_q-T_s+1:t_q}^{(s)}\), UNICON predicts
\(\hat V_q\in\mathbb{R}^{N_s\times C_s}\). Model weights remain fixed
throughout evaluation; no gradient-based parameter updates, system-specific
input or output layers or architectural changes are introduced.

\section{Data}

\subsection{Training corpus}

The training corpus comprises 20 data sources. Here, a source denotes a
dataset together with the geographical subset or dynamical regime from which
training examples are drawn; several sources can therefore originate from one
underlying data product. Each source retains its own graph, channel, sampling
rate and history length, together with four source-specific forecast horizons.
For each training task, one of these horizons is selected, and its
examples and query use the same horizon. Transformations and normalization
statistics are estimated separately for each source using only its permitted
training period, and the 20 sources are sampled with equal probability during
mixed-source training.

The 20-source mixture spans hydrology, traffic, power systems, weather, land,
ocean, soil, solar resources and human mobility. Four hydrology sources are
drawn from Caravan: CAMELS-US, CAMELS-BR, HYSETS and
CAMELS-GB\cite{kratzert2023caravan,newman2014camels,chagas2020camelsbr,arsenault2020hysets,coxon2020camelsgb}.
The two traffic sources are PEMS04 and
PEMS08\cite{guo2019astgcn,caltransPems}, and the two power-system sources are
EIA-930 and a PyPSA-Eur network populated with Open Power System Data time
series\cite{eia930,pudlEia930,horsch2018pypsaeur,opsd2020timeseries}. Eight
Earth-system sources span weather, land, ocean and soil: GLDAS, ERA5-Land,
WeatherBench2 ERA5, SMAP-L4, three GLORYS regions (North Pacific, Southern Ocean and Kuroshio), and the Copernicus Mediterranean Sea
reanalysis\cite{beaudoing2020gldas,munozsabater2021era5land,hersbach2020era5,rasp2024weatherbench2,reichle2025smapl4,lellouche2021glorys12,copernicusGlorys,copernicusMedsea}.
The remaining four sources comprise NASA POWER solar-resource fields over the
United States and Australia\cite{nasaPowerHourly}, and human-mobility
measurements from MTA Subway ridership and Citi Bike trip
histories\cite{mtaSubwayRidership,citibikeSystemData}.

\subsection{Evaluation datasets and splits}

The evaluation spans three levels of separation from the training corpus.
PEMS04 and CAMELS-US use unseen periods from sources included in training.
METR-LA, PEMS07, CAMELS-CL and LamaH-CE are unseen datasets from the
represented disciplines of traffic and
hydrology\cite{li2018dcrnn,caltransPems,alvarez2018camelscl,klingler2021lamah}.
AirQualityBench, WikiMaths and SuperMAG are systems from disciplines absent from training: global multi-pollutant air quality,Wikipedia mathematics page-view activity and ground magnetic-field perturbations,
respectively\cite{xu2026airqualitybench,openaq,rozemberczki2021pytorchgeoTemporal,gjerloev2012supermag}.

All model-input normalization, specialist training and contextual examples use
only data from each dataset's permitted pre-evaluation period. No query receives
its own target or any future observation. The nine evaluation systems comprise two unseen temporal splits
from training sources, four unseen datasets from represented disciplines and
three systems from disciplines absent from training.

\section{UNICON architecture}
\label{sec:unicon-architecture}
\subsection{Graph representation}

UNICON represents each system \(s\) as a graph
\(G_s=(\mathcal{V}_s,\mathcal{E}_s)\), whose \(N_s\) nodes denote the
system-specific entities at which variables are observed and predictions are
made. Existing relational structure is retained when available, whereas
gridded fields and spatially indexed observations are represented as graphs
over their valid grid cells, stations or regions. Each node carries \(C_s\)
channels, each corresponding to one system-specific variable observed over
time. The graph encodes relations among entities, whereas the channel axis
organizes the quantities associated with each entity. Systems may differ in
graph size, topology, edge attributes, the number and set of variables, units
and spatial resolution. This common graph representation does not require
systems to share a variable vocabulary. The graph specifies the relational
structure, whereas numerical examples express the system- and task-specific
predictive relation.

\subsection{Input representation}

For each prediction task, UNICON receives \(D\) key-value pair examples and a query, all defined over the same system graph. Each example pairs an
observed history (key) \(K_i\in\mathbb{R}^{T_s\times N_s\times C_s}\) with its corresponding future state (value) \(V_i\in\mathbb{R}^{N_s\times C_s}\) at the selected forecast horizon. The query provides an observed history
\(K_q\in\mathbb{R}^{T_s\times N_s\times C_s}\), while its corresponding future
state \(V_q\) is withheld. These inputs are organized as the
sequence
\[
  [K_1,V_1,\ldots,K_D,V_D,K_q].
\]
The system and task jointly determine the dimensions: \(T_s\) is the history
length, \(N_s=|\mathcal{V}_s|\) is the number of graph nodes and \(C_s\) is the
number of system-specific variables. Different systems therefore use the same
sequence construction while retaining their own histories, graphs and variable
sets.

Each key history is encoded independently at every node-variable location. A
temporal induced-token encoder first lifts the scalar sequence
\(K_{i,:,n,c}\) to the model dimension and adds a lookback-distance embedding.
It then applies Pooling by Multihead Attention (PMA), the Set Transformer
attention-pooling operation~\cite{lee2019set}. PMA uses a learned seed vector
to attend over the \(T_s\) lifted time steps and return a fixed-dimensional
token \(z^K_{i,n,c}\in\mathbb{R}^{d}\). The same temporal encoder is shared by
the contextual keys and the query key; values use a separate encoder of the
same form to produce \(z^V_{i,n,c}\in\mathbb{R}^{d}\). Each encoded key or value
is therefore a node-variable token array in
\(\mathbb{R}^{N_s\times C_s\times d}\).

The encoded arrays are assembled into the interleaved token tensor
\[
  [z^K_1,z^V_1,\ldots,z^K_D,z^V_D,z^K_q]\in
  \mathbb{R}^{(2D+1)\times N_s\times C_s\times d},
\]
where \(z^K_q\) is the query token array. A causal mask enforces the
autoregressive ordering along the sequence: the representation at each key
position can use the preceding key-value pairs but not its paired value or any
later element, while the final query can use all \(D\) contextual examples.

\subsection{Subgraph processing and edge aggregation}

\begin{figure}[!htbp]
\centering
\includegraphics[width=\linewidth]{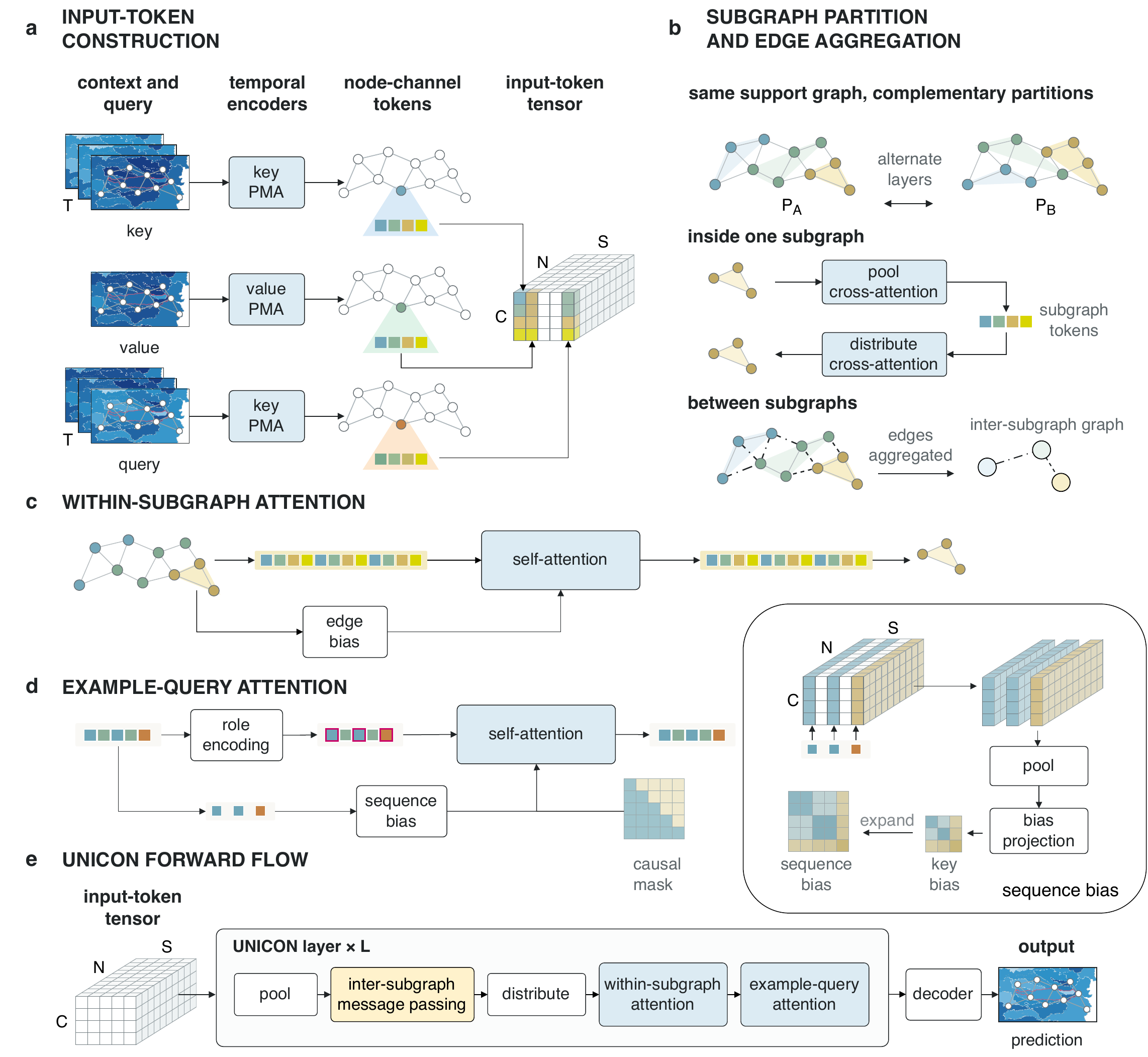}
\caption{\textbf{UNICON architecture.}
\textbf{a,} Input construction. Contextual examples pair observed histories
(keys) with future states (values), while the query contains an unpaired
history. PMA-based encoders map these inputs to node-channel representations.
\textbf{b,} Subgraph partition and edge aggregation. Two complementary
partitions define different node groups and alternate across layers. Graph
topology and edge attributes condition information exchange within and across
subgraphs.
\textbf{c,} Within-subgraph attention. Graph-conditioned self-attention updates
the local node-channel representations within each subgraph.
\textbf{d,} Example-query attention. Causal attention operates over the
interleaved key-value sequence at each node-channel location, allowing the
query to use the preceding contextual examples.
\textbf{e,} Layer stack and decoder. Each layer combines graph-conditioned
processing with example-query attention, and a shared decoder maps the
resulting representations to future-state predictions.}
\label{fig:unicon-architecture}
\end{figure}

UNICON processes the token tensor through a stack of graph-conditioned layers
(Extended Data Fig.~\suppref{fig:unicon-architecture}). Successive layers
alternate between the two complementary partitions \(P_s^A\) and \(P_s^B\).
Within each layer, tokens are processed according to the selected subgraphs,
while graph topology and edge attributes guide information exchange within and
across subgraphs. The resulting updates are returned to the node
representations and applied to every contextual example and the query.

\subsection{Within-subgraph attention}

Within each subgraph, graph-conditioned self-attention updates the local token
representations. The same parameterization is used across systems, allowing
the operation to accommodate different graph structures and variable sets.
The updated representations are then passed to the subsequent graph and
example-query processing layers.

\subsection{Example-query attention}

The meanings and predictive roles of system-specific variables are established through the contextual examples. After within-subgraph mixing, the tensor is reshaped to \(B\times(N_sC_s)\times(2D+1)\times d\), and example-query attention applies causal self-attention along the interleaved key--value and query sequence independently at each position--variable location. A variable remains aligned across every contextual key, its paired value and the query. The examples can therefore show how that variable, together with the cross-variable and graph-conditioned information already incorporated by within-subgraph attention, maps from past observations to a future state. Applying within-subgraph and example-query attention in each layer allows the model to learn from the demonstrated key--value pairs how these interactions bear on the query prediction.

The example-query attention layer also builds a content-dependent sequence bias from the key-history tokens. Each key is averaged over position--variable locations, transformed by a small multilayer perceptron and compared with the other keys through learned per-head projections. The resulting key-similarity matrix is expanded to the full interleaved sequence by assigning each value token to its paired key. This matrix supplies an additive bias to the attention logits and is combined with the causal mask. A learned role embedding distinguishes key and value tokens. A shared scalar decoder maps the final representations at all key positions to future-state predictions \(\hat V_1,\ldots,\hat V_D,\hat V_q\in\mathbb{R}^{N_s\times C_s}\).

\section{Training and retrieval}

\subsection{Training objective}

The training objective is mean squared error on normalized future states.
During mixed-source training, the decoder produces a prediction for every key and for the query in each sequence. These predictions are compared with the
corresponding values and the query's future state. For the reported
model, the prediction for the first key is excluded from the loss,
because no preceding example is available from which to infer the underlying dynamics. When a dataset provides structural observation masks, the loss is restricted to observed entries; otherwise, it is computed over all entries of the future state.

\subsection{Context retrieval}
\label{sec:supp-context-retrieval}

Examples are selected separately for each system. A candidate is eligible only
if its complete target is observable no later than the query issue time. During
mixed-source training, each source samples examples from precomputed
nearest-neighbor candidates. During evaluation, cosine similarity is computed
by flattening the normalized query and candidate keys over the configured
retrieval window, nodes and channels. Unless otherwise stated, evaluations use
the five eligible examples with the highest similarity to the query. For all
datasets except AirQualityBench, the retrieval window matches the model input
window. For AirQualityBench, retrieval uses only the final two hours of the
24-hour input to limit the dimensionality of each station-wise search vector.
For all evaluations, the eligible candidate set is fixed before evaluation. The
retrieved condition and the random, mismatched and noisy controls draw from the
same eligible candidates.

\section{Agent-assisted inference with frozen UNICON}
\label{sec:supp-agent-assisted-inference}

UNICON remains frozen throughout agent-assisted inference. For each task, a large language model (LLM) agent receives a task description and instructions, together with input observations and examples from the corresponding dataset. The agent can query the same frozen UNICON model multiple times and use the returned predictions to develop a task-specific prediction procedure. This procedure is then applied with frozen UNICON to generate the final predictions for the test data, without updating any model weights.

\subsection{Agent-assisted inference workflow}
The instructions supplied to each agent comprised a root instruction, a launch prompt, general skills and dataset-specific materials. The root instruction defined the objective and required output, while the general skills described how to inspect the task, compare alternative approaches and validate the output. Dataset-specific descriptions and skills varied across datasets and explained the data, evaluation metric, available observations and how to query frozen UNICON.

\begin{figure}[!htbp]
\centering
\includegraphics[width=\linewidth]{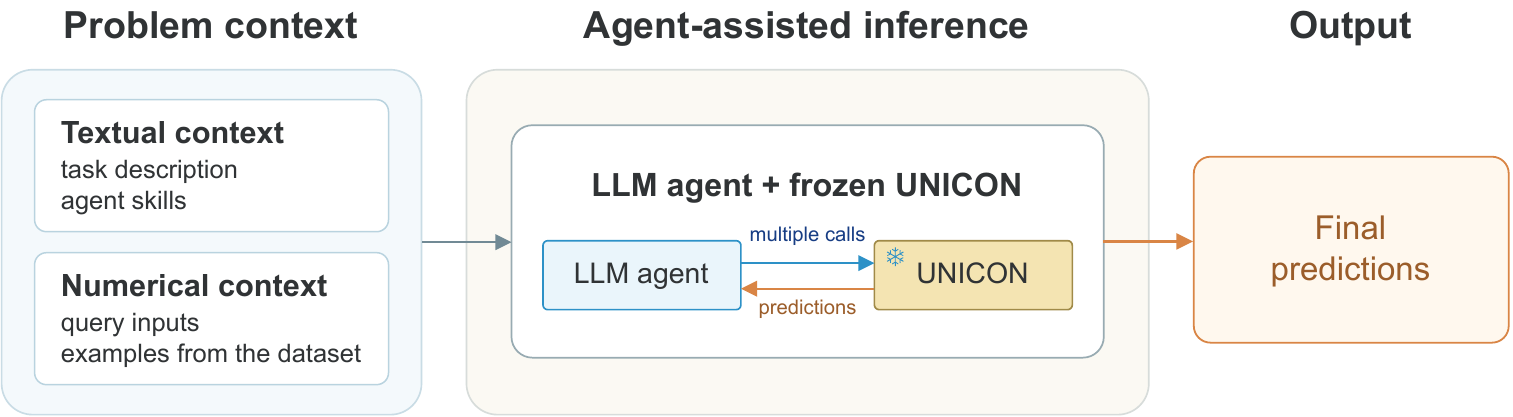}
\caption{\textbf{Agent-assisted inference with frozen UNICON.}
Textual context provides the task description and agent skills, while numerical
context provides input observations and examples from the dataset. A large
language model (LLM) agent can query the same frozen UNICON model multiple
times and use the returned predictions to develop a task-specific prediction
procedure. Applying this procedure with frozen UNICON generates the final
predictions for the test data. Supplementary
Table~\ref{tab:agent-adapters} summarizes how frozen UNICON was used for each
task.}
\label{fig:agent-context-interaction}
\end{figure}

Depending on the task, the agent compared, selected, combined or adjusted UNICON predictions, or retained the direct predictions unchanged. Extended Data Fig.~\suppref{fig:agent-context-interaction} shows the common workflow, whereas Supplementary Table~\ref{tab:agent-adapters} summarizes how frozen UNICON was used for each task.

\FloatBarrier

\subsection{Agent instructions and skills}

The excerpts below show the root instruction and four general skills shared
across datasets. We then show the PEMS04 launch prompt, task description and
PEMS04-specific skill as an example of the instructions written for an
individual task.

% BEGIN VERBATIM INSTRUCTION DISCLOSURE
\begin{agentinstructionbox}{Shared root instruction and general skills}

\agentexcerptheading{Root instruction}

\agentexcerptlabel{Task information}

Recover the objective, data and prediction semantics, metric, available
observations, frozen-model tools, feedback schedule, budget and required
submission.

\agentexcerptlabel{Objective}

Use the supplied observations and frozen numerical-model tools to design, test
and package a complete executable inference policy. The numerical model is
frozen: do not update its weights or train a replacement neural specialist. The
concrete policy is your decision and should follow from the researcher-crafted
problem prior and observed examples.

\agentexcerptlabel{Working boundary}

\begin{itemize}
\item Use only paths, data and tools declared by the problem.
\item Respect the observation and feedback order exposed by the problem
  interface.
\end{itemize}

\agentexcerptlabel{Completion}

Preserve a completion-safe fallback when the problem defines one. Reserve
enough budget to validate the packaged policy, satisfy every required
submission field and finish complete coverage.

\agentexcerptdivider
\agentexcerptheading{General skill: Inspect the problem first}

Treat the problem statement as researcher-authored prior knowledge, not as a
form whose headings replace scientific explanation. Inspect any supplied
resources and interfaces before broad experimentation.

\agentexcerptlabel{The requested problem map contained:}

\begin{itemize}
\item objective, query semantics, prediction shape and physical or normalized
  units;
\item exact metric and its independent aggregation units;
\item available observations, phase boundaries and any feedback schedule;
\item frozen-model and retrieval tools, including their expected cost;
\item direct reference or completion-safe fallback when declared;
\item commitment rule, formal execution state and completion condition;
\item required policy function, files, records and validation checks.
\end{itemize}

Separate facts and researcher-supplied priors from hypotheses you introduce
yourself. If supplied files disagree about an interface, stop and resolve the
inconsistency rather than silently redefining the task.

\agentexcerptdivider
\agentexcerptheading{General skill: Design an inference policy}

Treat frozen-model calls, supplied retrieval and task-specific calculations as
components of an executable policy. Begin from a declared direct or
completion-safe reference when one exists, then test materially different
hypotheses rather than renamed or numerically equivalent variants.

Measure prediction-space diversity, matched-evidence benefit, failure behavior
and cost. Prefer the smallest inspectable rule whose benefit persists across
the provided validation structure. More complex selectors, mixtures or
query-dependent rules must earn their complexity outside the rows used to fit
them.

When the supplied frozen-model interface exposes deterministic,
structure-preserving configurations, treat them as optional test-time calls
rather than implicit model updates. Compare only the bounded configurations
declared by the problem, preserve the canonical call as fallback, and commit
the selected configuration before formal inference.

Keep every learned coefficient and required asset inside the final policy. Make
the policy deterministic, finite and reconstructible from the current query,
supplied observations and declared tools. Do not rely on unavailable
cross-query state or model-weight updates.

\agentexcerptdivider
\agentexcerptheading{General skill: Evaluate on observed examples}

Follow the observation schedule exposed by the problem. When the supplied
examples support separate development, selection and confirmation phases,
assign those roles before broad search. When feedback arrives sequentially,
update only from observations already returned by the interface.

Compare candidates on the same queries, masks and metric. Treat the problem's
independent units, not the largest tensor dimension, as the units for
validation and complexity control. Examine paired behavior, temporal or
structural folds, tail failures and support coverage when they are relevant.

For ordered problems, preserve the order in the diagnosis rather than treating
held-out units as exchangeable replicates. Report rolling-origin or
successive-block effects through the boundary nearest the unresolved queries. A
benefit that steadily decays, a coefficient that drifts, or a correction
supported mainly by early blocks is evidence of extrapolation risk even when
the pooled held-out mean remains positive. Let that evidence determine whether
to refit on a more relevant window, attenuate toward the reference, introduce a
cross-fitted state rule, or reject the added correction.

Keep the selection rule explicit and stable. A negative confirmation remains
part of the experiment record; do not replace a committed policy through an
undeclared second selection stage.

Treat the direct frozen-model prediction as a real candidate, not merely a weak
baseline. Commit an intervention only when its benefit is supported by the
problem's independent units and persists through the most relevant
chronological or structural validation boundaries. If the sign, scale or
coverage of the benefit is unstable, shrink toward the direct candidate or
submit it unchanged. The objective is an evidence-backed intervention, not a
correction at any cost.

\agentexcerptdivider
\agentexcerptheading{General skill: Allocate compute and complete the task}

Measure data loading, one model call and one metric pass before committing the
full budget. Reserve time for complete coverage, policy import or execution
checks, required records and a safe fallback.

Use the remaining budget on a small portfolio of prediction-distinct
hypotheses and matched comparisons. Prune redundant or repeatedly unstable
directions, reallocate to informative disagreements or alternate folds, and
stop a family when visible evidence shows saturation. More calls are not useful
when they cannot affect the final policy.

When labeled public examples are abundant and measured calls are cheap, use a
progressive sample-size ladder before declaring saturation. A sparse regime or
quantile screen can prune hypotheses, but it does not by itself establish
coefficient, selector or held-out-effect stability. Expand matched coverage
until the selected policy and its validation effect are stable under
subsampling, or record the concrete time, redundancy or instability evidence
that stops expansion.

Track elapsed time, calls, candidate identities, evidence roles, retention
decisions and remaining completion margin. Degrade gracefully as the deadline
approaches. Finish only after every required submission file exists, the policy
satisfies the declared interface, outputs are finite and the completion-safe
path has been exercised.

\end{agentinstructionbox}

\begin{agentinstructionbox}{PEMS04-specific instructions and skill}

\agentexcerptheading{Launch prompt}

Use the two supplied chronological public blocks deliberately, compare
materially different low-complexity policies, and spend the remaining time
closing a complete executable submission rather than stopping after the first
positive number. Treat exact direct UNICON as a genuine candidate: intervene
only when the supplied validation structure supports transfer, otherwise
submit the direct policy unchanged.

\agentexcerptdivider
\agentexcerptheading{Problem-specific example: PEMS04 traffic forecasting}

\agentexcerptlabel{Objective}

Improve the supplied frozen-UNICON forecast for traffic flow at 12 successive
five-minute leads. PEMS04 was represented in the model's training corpus, while
the observations supplied here form a separate chronological forecasting
exercise. The headline measure is the mean of the 12 original-unit MAEs after
the declared zero mask.

\agentexcerptlabel{Scientific context}

Each query follows 12 observed frames on a 307-sensor network. Candidate zero
is the exact direct UNICON forecast and must remain the unconditional fallback.
The supplied query state retains the complete 12-frame physical history and
mask, clock phase, daily and weekly observations, and the declared road graph.
A useful rule may calibrate horizons, exploit stable temporal or network
structure, or combine genuinely complementary frozen-model calls, but
complexity must improve more than one chronological block.

\agentexcerptlabel{Boundaries}

\begin{itemize}
\item Keep the numerical-model weights unchanged and do not train a replacement
  neural forecaster.
\item Use only the supplied observations, candidate forecasts and query
  features.
\item Make every forecast depend only on its own query record; the packaged rule
  must pass singleton replay without using later query inputs.
\item Keep prediction logic deterministic, reconstructable and complete for
  every query.
\item Preserve the exact direct forecast whenever optional information is
  unsupported or malformed.
\end{itemize}

\agentexcerptlabel{Completion}

The prediction must match the direct candidate's \texttt{[time, node, horizon]}
shape and be finite. Validate the packaged policy on both supplied
chronological blocks.

\agentexcerptdivider
\agentexcerptheading{Problem-specific skill: Work with traffic evidence}

Prefer shared, horizon-level or otherwise strongly regularized corrections.
Check chronological stability, fallback coverage and whether gains survive
different traffic regimes. Do not fit an independent answer for every cell.
Leave enough time to validate the final files.

The named features expose all 12 issue-time frames, day/week phase and matching
daily/weekly observations. Treat graph rows as fixed network context. Begin
with low-dimensional summaries, but derive them from this complete state and
record what was used rather than assuming the convenience projection is the
forecasting problem.

Keep candidates, folds and scoring identical. If the richer view does not
improve transfer, retain the compact or direct rule and record that public
result as the sufficiency evidence.

When a fixed candidate improves one chronological block but weakens later,
treat that as a transfer problem rather than adding another global offset.
Compare it with at least one small, rolling-origin residual rule whose inputs
are available for the unresolved query itself, such as candidate disagreements
and observed current, lagged, recent, daily or weekly values. Fit and
standardize on earlier public rows, use strong shrinkage or another explicit
low-dimensional constraint, and choose complexity on later public rows. Package
every coefficient and preprocessing constant needed for deterministic replay.

\end{agentinstructionbox}
% END VERBATIM INSTRUCTION DISCLOSURE

\subsection{Prediction procedures developed by the agent}

For most tasks, the agent submitted a task-specific prediction procedure rather than predictions for the test data. The procedure specified how to query frozen UNICON and how to select, combine or adjust the returned predictions.

The WikiMaths example below illustrates a prediction procedure developed by
the agent. It starts from the direct UNICON prediction and combines it with
three additional predictions generated from different sets of examples. Their
disagreement determines the scale of the adjustment.

\begin{agentprocedurebox}{WikiMaths prediction procedure}

\begin{agentproceduresteps}
\textbf{Input} & Input observations for one test case, denoted \(q\). \\[2pt]
\agentprocedurestep{1} &
  \(\hat{\mathbf y}^{(0)}\gets\operatorname{UNICON}(q)\) using five retrieved
  contextual examples. \\[2pt]
\agentprocedurestep{2a} &
  \(\hat{\mathbf y}^{(1)}\gets\operatorname{UNICON}(q)\) using eight weekly
  seasonal examples. \\
\agentprocedurestep{2b} &
  \(\hat{\mathbf y}^{(2)}\gets\operatorname{UNICON}(q)\) using eight
  scale-matched examples. \\
\agentprocedurestep{2c} &
  \(\hat{\mathbf y}^{(3)}\gets\operatorname{UNICON}(q)\) using twelve
  retrieved contextual examples. \\[2pt]
\agentprocedurestep{3} &
  Set \(a(q)\in[0,1]\) by applying the fixed scaling rule to the disagreement among
  \(\hat{\mathbf y}^{(1)}\), \(\hat{\mathbf y}^{(2)}\) and
  \(\hat{\mathbf y}^{(3)}\). \\[2pt]
\textbf{Output} &
  \(\displaystyle
  \hat{\mathbf y}\gets\hat{\mathbf y}^{(0)}+
  a(q)\sum_{j=1}^{3}\beta_j
  \left(\hat{\mathbf y}^{(j)}-\hat{\mathbf y}^{(0)}\right)\).
\end{agentproceduresteps}

\medskip
\noindent The scaling rule for \(a(q)\) and the coefficients
\(\boldsymbol{\beta}=(0.4065,0.3068,0.4142)\) were fixed before test
evaluation; the displayed coefficients are rounded to four decimal places.

\end{agentprocedurebox}

\begingroup
\small
\setlength{\tabcolsep}{4pt}
\setlength{\LTcapwidth}{\linewidth}
\begin{longtable}{@{}>{\raggedright\arraybackslash}p{0.22\linewidth}>{\raggedright\arraybackslash}p{0.72\linewidth}@{}}
\caption{\textbf{How agent-assisted predictions were produced.}
Each row summarizes the
prediction procedure developed for one task by the agent.}\label{tab:agent-adapters}\\
\toprule
Task & Prediction procedure \\
\midrule
\endfirsthead
\toprule
Task & Prediction procedure \\
\midrule
\endhead
\endfoot
\bottomrule
\endlastfoot

\multicolumn{2}{@{}l}{\textbf{Unseen periods from training sources}} \\*
PEMS04 & Combined UNICON predictions from different retrieved examples and applied a horizon-specific correction using the preceding 12 traffic observations and time of day and week. \\
CAMELS-US & Adjusted each basin's direct UNICON prediction towards its current observed streamflow using a basin-specific weight estimated from earlier examples, with outputs constrained to be nonnegative. \\

\addlinespace
\multicolumn{2}{@{}l}{\textbf{Unseen datasets from represented disciplines}} \\*
PEMS07 & Blended UNICON predictions from different retrieved examples, then applied a regularized horizon-specific correction using recent traffic history, time of day and week, and neighbouring-sensor flow. \\
METR-LA & Used the UNICON prediction based on ten retrieved examples when valid, otherwise retaining the direct UNICON prediction. \\
CAMELS-CL & Adjusted each basin's direct UNICON prediction towards its current observed streamflow using basin-specific weights regularized towards a shared value. \\
LamaH-CE & Retained the direct UNICON prediction because adjustments based on current, recent or longer streamflow histories did not transfer consistently across earlier evaluation periods. \\

\addlinespace
\multicolumn{2}{@{}l}{\textbf{Systems from disciplines absent from training}} \\*
AirQualityBench & Adjusted direct UNICON predictions using pollutant observations from the same hour on the previous day and the most recent hour, with coefficients fitted on chronological 2024 data. \\
WikiMaths & Made four UNICON calls using standard, weekly seasonal, scale-matched and deeper retrieval examples, then combined their predictions with a weight determined by their disagreement. \\
SuperMAG & Combined predictions from five and twenty fixed pre-event retrieved examples with the current query's last observed field, using weights selected from pre-event context only. \\

\end{longtable}
\endgroup

\section{Evaluation}
\label{sec:supp-evaluation}

\subsection{UNICON evaluation protocols}

This subsection defines the datasets, input windows, forecast horizons, metrics
and reference methods used to evaluate frozen UNICON. The evaluation systems
span three levels of separation from the training corpus. PEMS04 and CAMELS-US
contain unseen periods from sources included in training. PEMS07, METR-LA,
CAMELS-CL and LamaH-CE are unseen datasets from disciplines represented in
training. AirQualityBench, SuperMAG and WikiMaths are from disciplines absent
from training. Within each comparison, all methods are evaluated on the same
cases and targets using the same scoring procedure. Scores are not pooled
across tasks that use different metrics or units.

\textbf{Traffic.} METR-LA, PEMS04 and PEMS07 are evaluated at 5-min
cadence. Each query contains a 12-frame observed history. Forecast leads
\(\Delta=1,\ldots,12\), corresponding to 5-60 min, are predicted directly
and independently without autoregressive rollout. The complete observed history is used for retrieval. PEMS04 follows the ASTGCN 60/20/20 split, METR-LA follows the DCRNN 70/10/20 split and PEMS07 uses the public
ASTGCN/STAEformer-style test segment. Predictions and targets are transformed back to their original units before scoring. The evaluated variable is speed for METR-LA and flow for PEMS04 and PEMS07. Ground-truth values equal to zero are treated as missing under the public traffic protocol. MAE is computed separately at each lead and then averaged across all 12 leads.

Traffic specialists are trained using the official STAEformer implementation for PEMS04 and PEMS07\cite{liu2023staeformer} and the official BasicTS
D\(^{2}\)STGNN implementation for METR-LA\cite{shao2022d2stgnn}. Each
specialist retains the training and validation splits of its reference
implementation. Specialists and UNICON are evaluated on the same cases and
targets, comprising 3,387 PEMS04, 6,843 METR-LA and 5,633 PEMS07 query windows, each with 12 forecast leads. Checkpoints are selected using validation performance only. For each dataset, the specialist reference is the mean score across three independently initialized runs; no run is selected using test
performance.

\textbf{Hydrology.} CAMELS-US, CAMELS-CL and LamaH-CE are daily datasets
accessed through the Caravan v1.5 archive, with 531, 462 and 855 retained
basins, respectively. Each query contains a 30-day observed history and
predicts streamflow one day ahead. The same 30-day history is used for
retrieval, and no autoregressive rollout is applied. Evaluation covers the
five water years from 1 October 2003 to 30 September 2008, with examples
restricted to dates ending no later than 30 September 2003. Each input contains
14 ERA5-Land forcing variables and streamflow, but only streamflow is scored.
Predictions are denormalized to physical streamflow units. Missing observations
are excluded separately for each basin. Basins with fewer than two valid
targets or zero variance in observed streamflow are omitted from aggregation.
Median basin NSE is the reported aggregate metric. Absolute error in physical
units and catchment-mean MAE are used for the LamaH-CE case-study visualization
and flow-regime analyses.

\textbf{Air quality.} AirQualityBench contains hourly measurements of
PM\(_{2.5}\), PM\(_{10}\), NO\(_2\), O\(_3\), SO\(_2\) and CO at 3,720
stations. Candidate examples are drawn from 2021--2023, 2024 is used for
validation and the complete 2025 test year is used for evaluation. The model
receives a 24-hour observed history. All reported comparisons evaluate the
first six direct hourly leads, with issue times spaced six hours apart.
Predictions are denormalized before scoring. These comparisons follow the
released D\(^{2}\)STGNN scoring procedure. The 1,456 chronological issue times
are retained in their released order and divided without shuffling into 364
batches of four. Within each batch, masked MAE in physical units is computed
over all issue times, leads, stations and pollutants. The resulting batch
scores are averaged equally. All reported AirQualityBench results use this
scoring procedure and the released observation mask.

\textbf{Web activity.} WikiMaths comprises 731 daily observations of page-view
activity on a graph of 1,068 mathematics-related Wikipedia pages. The first
80\% of the series is available for specialist training and example selection,
the next 10\% is used for validation and the final 10\% for testing. Each query
and retrieval vector contains an eight-day observed history, and the future
state is the following day. No log transform is applied. Inputs are normalized
per node using statistics computed from the initial 80\% period. Each
validation and test date is evaluated as one full-graph query. Evaluation
follows the public PyTorch Geometric protocol, with mean squared error reported
over all finite node-time entries in its per-node standardized space. The
specialist reference is LAMP\cite{castelli2026adaptive}, trained using its
official 32-setting grid and selected using validation MSE only. We retrain
LAMP using target-aligned splits containing 576 training targets, 73 validation
targets and 73 test targets. The scorecard compares kNN, LAMP and UNICON on the
common test targets in the public protocol's full-series per-node standardized
space. Fig.~\mainref{fig:prediction-views}b instead compares LAMP and UNICON on
the same targets after the same training-period per-node standardization.

\textbf{Space weather.} We forecast ground magnetic-field perturbations 30 min
ahead during the May 2024 Gannon storm using the northward and eastward
components measured by SuperMAG and distributed in the GeoDGP-associated Deep
Blue Data release\cite{gjerloev2012supermag,chen2024geodgpdata}.
We also cite the requested technical descriptions for contributing stations
from EMMA, IMAGE, MACCS, McMAC, MAGDAS/210, CARISMA, PENGUIn (AAL-PIP) and
INTERMAGNET\cite{lichtenberger2013plasmon,tanskanen2009image,
engebretson1995maccs,chi2013mcmac,yumoto2001cpmn,mann2008carisma,
clauer2014aalpip,love2013intermagnet}. Each query contains the preceding 2 h of
measurements on a fixed spatial graph. Earlier
storms from 2010--2015 provide examples available before the event, and the
Gannon interval contains 619 forecast times. Locations without valid
measurements are excluded from scoring rather than filled by spatial
interpolation.

Frozen UNICON and weighted kNN receive the same five examples for each query.
The Graph WaveNet specialist reference\cite{wu2019graphwavenet}
was trained only on the earlier storms and is not updated during Gannon.
Predictions are converted to nT and scored using root mean squared error over
the two horizontal magnetic-field components at observed locations. This is a
held-out evaluation of the May 2024 Gannon storm.

\subsection{Evaluation protocol for agent-assisted inference}

For each task, we apply the submitted procedure to the test data to generate
the agent-assisted predictions. Direct and agent-assisted inference use the
same frozen UNICON checkpoint, evaluation cases, preprocessing and scoring
procedure.

Each procedure is fixed before test evaluation and applied without revision to
the complete test set. Test targets and evaluation scores remain unavailable to
the agent throughout this process. Each SuperMAG prediction uses only its own
24-frame input history and the fixed pre-event context pool; test queries do
not update later predictions.

\section{Additional analyses and controls}

\subsection{Dataset-specific analyses}

\subsubsection{LamaH-CE flow-regime, peak-window and forecast-horizon analyses}

We aligned the one-day-ahead UNICON and supervised-LSTM predictions\cite{zhang2024eddlstm} over the 855 LamaH-CE catchments and their common test dates. Flow regimes were defined separately for each catchment from the empirical distribution of its observed test-period streamflow. Within each percentile interval, MAE was first computed for each catchment and then averaged across catchments, giving every catchment equal weight. Ratios compare this catchment-mean MAE for UNICON with that of the LSTM specialist. Uncertainty was estimated with 50,000 paired bootstrap replicates that resampled catchments and retained the complete aligned time series of both models.

High-flow events were defined independently of either model as the maximum observed streamflow in each complete October--September water year. We computed MAE on the peak day and within symmetric windows extending 1, 3, 7 and 15 days on either side. Events with incomplete windows or missing aligned observations were excluded. Event errors were averaged within each catchment before aggregation, and 95\% intervals for the UNICON-to-LSTM MAE ratio were obtained by the same paired catchment bootstrap. The low-flow percentile analysis describes predictive error under low observed flows.

The forecast-horizon comparison used direct predictions at 1, 3, 7 and 14 days for 855 catchments and 1,813 common target dates from 15 October 2003 through 30 September 2008. Frozen UNICON used the same checkpoint at every horizon, with contextual histories paired with states at the requested horizon. A separate supervised LSTM was trained directly for each horizon using the same 30-day histories, catchment set, training and validation periods and streamflow normalization; neither method used recursive rollout or future meteorological forcing. Both methods were evaluated on the same aligned catchments, target dates and observations. NSE was computed per catchment and summarized by the median over 836 catchments with sufficient observations. Median intervals and paired differences used 50,000 bootstrap replicates that resampled catchments.

\begin{figure}[!htbp]
\centering
\includegraphics[width=\linewidth]{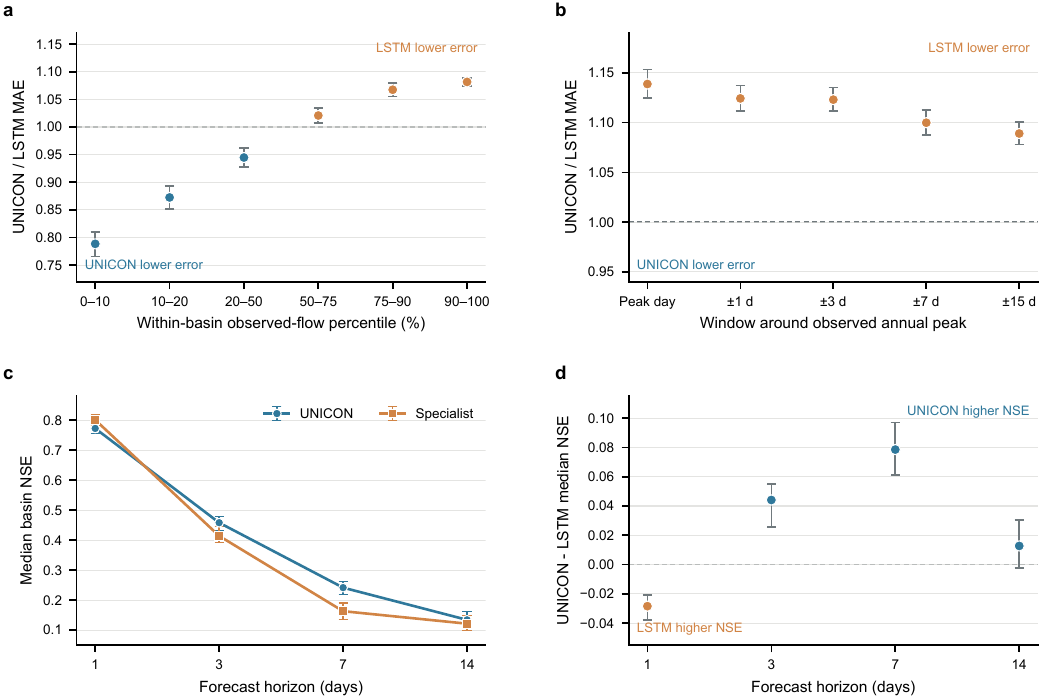}
\caption{\textbf{Flow- and horizon-dependent comparison of UNICON and a hydrological specialist.}
\textbf{a,} Ratio of one-day-ahead MAE for frozen UNICON to that of the supervised LSTM across within-catchment observed-flow percentiles. Percentile intervals are defined separately for each catchment.
\textbf{b,} The same ratio on the observed annual peak day and over symmetric windows around that peak. In \textbf{a,b}, points are ratios of catchment-mean MAE and bars are paired 95\% bootstrap intervals obtained by resampling catchments; ratios below one favour UNICON.
\textbf{c,} Median basin NSE for direct forecasts from frozen UNICON and separately trained horizon-specific LSTM specialists at 1, 3, 7 and 14 days. Bars show 95\% catchment-bootstrap intervals.
\textbf{d,} Paired difference in median basin NSE between UNICON and the LSTM specialist. Positive values favour UNICON; bars show paired 95\% catchment-bootstrap intervals and the dashed line denotes equal median NSE. All horizons use the same 1,813 target dates and 855 ordered catchments.}
\label{fig:lamah-flow-regimes}
\end{figure}

\subsubsection{Paired WikiMaths uncertainty and activity analysis}

We compared LAMP and UNICON on the same 73 WikiMaths target dates. Daily MSE and MAE were computed in the public full-series per-node standardized space. UNICON predictions are expressed in the training-period per-node standardized space used at inference; they were converted to the public space with the per-node affine relation recovered from the shared targets.

Uncertainty in the aggregate MSE difference was estimated by paired circular block bootstrap over dates. The same sampled date blocks were applied to both models, preserving their paired errors; the primary analysis used 50,000 replicates and a seven-day block. Sensitivity analyses used block lengths of 1, 3, 5, 7, 10 and 14 days. Graph-wide activity was defined independently of either model as the spatial mean absolute observed state in training-period per-node standardized space. High-activity dates were the top quintile of this quantity, and the date shown in Fig.~\mainref{fig:prediction-views}b was its maximum. Association between activity and the daily MSE difference was summarized by Spearman correlation; a two-sided $P$ value was obtained from all circular temporal shifts of one series relative to the other.

\begin{figure}[!htbp]
\centering
\includegraphics[width=\linewidth]{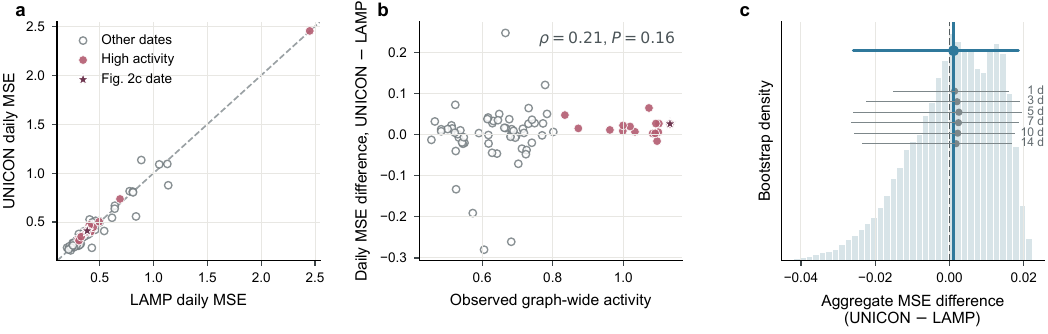}
\caption{\textbf{Paired comparison of UNICON and a WikiMaths specialist.}
\textbf{a,} Daily official MSE for the LAMP specialist and frozen UNICON over 73 aligned test dates. The dashed diagonal denotes equal error; filled points are the observation-defined top activity quintile and the star is the day shown in Fig.~\mainref{fig:prediction-views}b.
\textbf{b,} Daily MSE difference against graph-wide observed activity. Positive values favour LAMP. The annotation gives Spearman correlation and a two-sided circular-shift $P$ value.
\textbf{c,} Paired circular block-bootstrap distribution of the aggregate MSE difference using a seven-day block and 50,000 replicates. The blue point and thick interval show the observed difference and 95\% interval; grey intervals show sensitivity to block lengths of 1, 3, 5, 7, 10 and 14 days. The vertical dashed line denotes equal aggregate MSE.}
\label{fig:wikimaths-uncertainty}
\end{figure}

\subsubsection{WikiMaths forecast-horizon comparison}

We compared direct UNICON forecasts with autoregressive LAMP rollouts at forecast horizons of 1, 2, 3, 5, 7, 10 and 14 days. Both models were evaluated from the same eight-day observed histories at 60 common forecast origins. LAMP was applied recursively beyond the first day, whereas the frozen UNICON checkpoint predicted each requested horizon directly through its contextual interface. Evaluation used the same public full-series per-node standardized space. Uncertainty in the paired MSE difference was estimated with 10,000 circular block-bootstrap replicates over forecast origins, using a seven-day block.

\begin{figure}[!htbp]
\centering
\includegraphics[width=\linewidth]{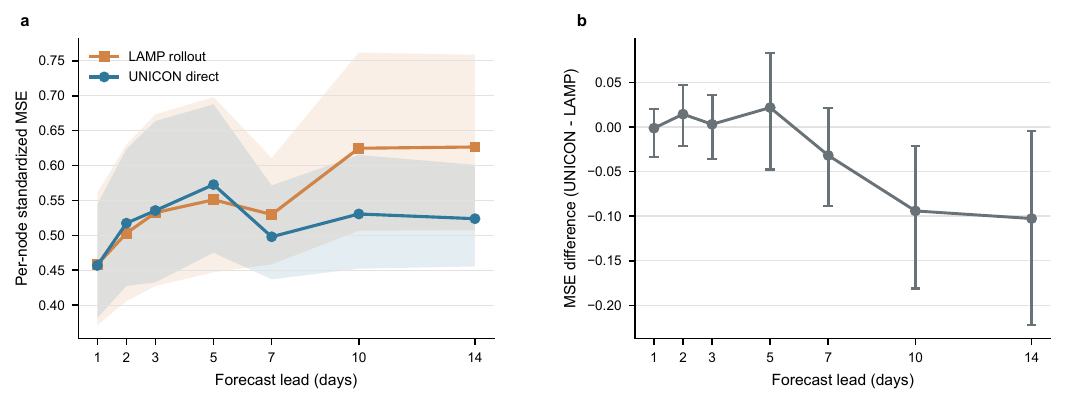}
\caption{\textbf{WikiMaths forecasts across forecast horizons.}
\textbf{a,} Per-node standardized MSE for autoregressive LAMP rollouts and direct UNICON forecasts over 60 common forecast origins. Shading denotes 95\% block-bootstrap intervals.
\textbf{b,} Paired MSE difference between UNICON and LAMP at each forecast horizon. Negative values favour UNICON; bars show 95\% block-bootstrap intervals and the dashed line denotes equal MSE.}
\label{fig:wikimaths-leadtime}
\end{figure}

\subsection{Contextual-example controls}

Contextual examples are selected using the retrieval procedure described in
Supplementary Material Section~\suppref{sec:supp-context-retrieval}. Each
example pairs an observed history with its corresponding future state, and
\(D\) denotes the number of examples given to the model. The retrieved
condition uses the five most similar examples, whereas the no-context
condition supplies none (\(D=0\)).

For every query, we generate and cache a fixed random ordering of up to 50
examples. For the random control, we use the first
\(D\in\{1,2,5,10,20,50\}\) examples in this ordering, so only the number of
examples changes. For retrieved examples, we use the first
\(D=1,\ldots,5\) examples in the similarity ranking.

The mismatched condition begins with five random examples and permutes their
future states so that none remains paired with its original history. The noisy
condition uses the same number of examples and the same observation masks, but
replaces the observed numerical values after normalization with values drawn
from a standard normal distribution. These controls compare retrieved
examples with no context, random examples, mismatched history--future pairs
and corrupted numerical values. Within each dataset, all conditions use the
same evaluation cases, masks, normalization, scoring procedure and query
order.

We express each dataset's reported metric as a lower-is-better error and define
normalized context gain as
\[
G(D)=\frac{E_0-E(D)}
           {E_0-E_{\mathrm{ret},5}},
\]
where \(E_0\) is the no-context error, \(E(D)\) is the error obtained with
\(D\) examples and \(E_{\mathrm{ret},5}\) is the error obtained with five
retrieved examples. Thus, \(G=0\) denotes no improvement over no context,
\(G=1\) matches the performance obtained with five retrieved examples and
negative values indicate worse performance than no context.

\begin{figure}[!htbp]
\centering
\includegraphics[width=\linewidth]{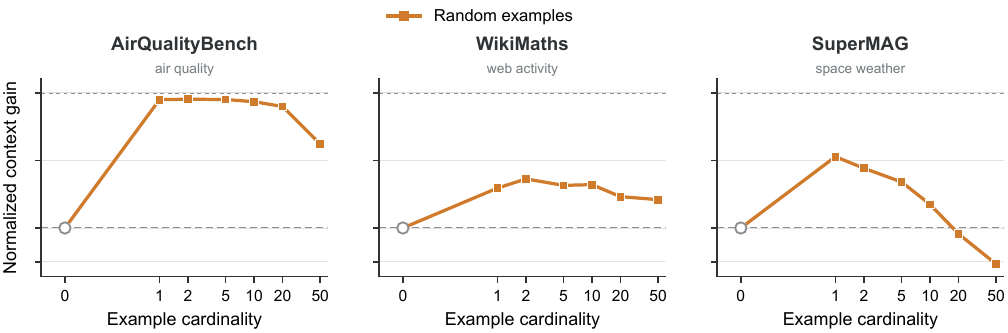}
\caption{\textbf{Random-example cardinality.}
Normalized context gain as the number of random examples increases for the
three systems from disciplines absent from the training corpus. The open
marker at \(D=0\) denotes no context. Horizontal dashed lines mark
no-context performance (zero) and the performance obtained with five
retrieved examples (one). The horizontal axis uses a symmetric logarithmic
scale to show cardinalities from zero to 50. Even without similarity-based
selection, one or two correctly paired random examples provide useful context
across all three systems, showing that a small number of numerical examples
can already convey substantial task information.}
\label{fig:random-cardinality}
\end{figure}

CAMELS-CL error is one minus median basin-wise NSE; traffic uses average MAE
over the 12 forecast horizons; AirQualityBench uses the six-hour output-window
MAE defined above; WikiMaths uses per-node standardized MSE; and SuperMAG uses
pooled horizontal-vector RMSE in nT. Figure~\ref{fig:example-controls}c shows
how performance changes with the number of retrieved examples. Extended Data
Fig.~\suppref{fig:random-cardinality} reports the corresponding results for
random examples.Even without similarity-based selection, one or two
correctly paired examples provide useful context across all three systems.

We also examine how adding one retrieved example affects individual queries
in AirQualityBench, WikiMaths and SuperMAG, whose disciplines were absent from
training. Let \(e_{0,q}\) and \(e_{1,q}\) denote the error of query \(q\) with
no context and with one retrieved example, respectively, and let \(E_0\)
denote the dataset-level no-context error. The query-level error reduction is
\[
\Delta_q=\frac{e_{0,q}-e_{1,q}}{E_0}.
\]
Positive values indicate improvement after adding one example. The shared
dataset-level denominator places query-level changes from different datasets
on comparable scales; it is independent of the five-example reference used
in \(G(D)\).

For each AirQualityBench issue time, error is computed jointly over all valid
station--pollutant targets in the first six hourly leads. The dataset-level
no-context error uses the same grouping of consecutive issue times into
batches of four as the reported AirQualityBench score. WikiMaths computes
per-node standardized MSE for each of 73 test dates, and SuperMAG computes
horizontal-vector RMSE for each of 619 Gannon-storm issue times. All queries
are retained. The fraction improved is the proportion for which
\(\Delta_q>0\).

\subsection{Corpus-diversity experiments}

To examine how training-corpus diversity affects cross-disciplinary
generalization, we trained models of the same size and with the same total
training budget on corpora that differed in the number and types of sources
included. Models trained on one source show the degree of specialization that
can be achieved for that system. Comparing them with models trained on more
diverse corpora shows how corpus diversity affects generalization to unseen
systems and whether performance is maintained on systems represented during
training.

\begin{figure}[!htbp]
\centering
\includegraphics[width=0.82\linewidth]{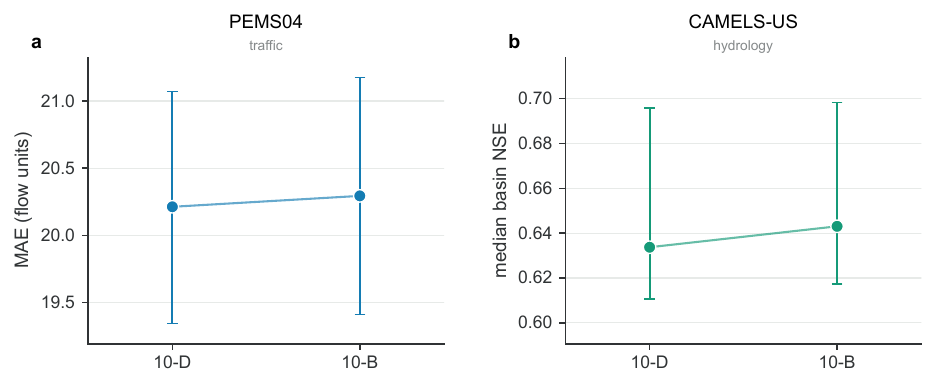}
\caption{\textbf{Comparison of two ten-source training corpora.}
\textbf{a,} PEMS04 MAE for models trained on the ten-source depth and breadth
corpora; lower is better.
\textbf{b,} CAMELS-US median basin NSE for the same models; higher is better.
Points show performance estimates and bars show 95\% bootstrap intervals
computed as in Fig.~\mainref{fig:source-diversity}b.
Both corpora include PEMS04, CAMELS-US, EIA-930, GLDAS and North Pacific
GLORYS with equal weight. The depth corpus (10-D) additionally includes
PEMS08, CAMELS-BR, PyPSA-Eur, the Copernicus Mediterranean Sea reanalysis and
ERA5-Land, adding further sources from the same five disciplines. The breadth
corpus (10-B) instead includes WeatherBench2 ERA5, Australia Solar, MTA Subway
ridership, Citi Bike and SMAP-L4, extending training to a wider range of
system types. Because the two corpora contain the same number of sources and
use the same total training budget, this comparison focuses on which sources
are included.}
\label{fig:ten-source-allocation}
\end{figure}

For the ten-source setting, we compare two ways of selecting the training
sources. The depth corpus adds sources from disciplines already represented
in the smaller training corpora. The breadth corpus instead includes a wider
range of system types while keeping the number of sources and total training
budget unchanged. The complete source lists are shown in
\hyperref[fig:ten-source-allocation]{Extended Data
Fig.~\suppref{fig:ten-source-allocation}}. The twenty-source corpus is the
most diverse training condition considered.

For Fig.~\mainref{fig:source-diversity}b, every checkpoint is evaluated on the
complete PEMS04 and CAMELS-US test sets using the same five retrieved examples
for each query. For models trained only on PEMS04 or CAMELS-US, we record
performance throughout training. Models trained on mixed corpora are evaluated
at their final checkpoints. The main figure shows the result for the
ten-source depth corpus, and the corresponding result for the breadth corpus
is reported in
\hyperref[fig:ten-source-allocation]{Extended Data
Fig.~\suppref{fig:ten-source-allocation}}.

The horizontal axis in Fig.~\mainref{fig:source-diversity}b estimates the
training compute allocated to PEMS04 or CAMELS-US. We express the cumulative
compute at training step \(t\) as
\[
\widehat C_{t,m}=\frac{t}{T_m}C_{\mathrm{total}},
\]
where \(C_{\mathrm{total}}=9.6\times10^{17}\) floating-point operations
(FLOPs) is the common training budget and \(T_m\) is the number of steps
required for corpus \(m\) to reach this budget. Because the computational cost
of a training step varies with corpus composition, \(T_m\) differs between
runs. The values used to construct the horizontal axis are provided in the
source data for Fig.~\mainref{fig:source-diversity}.

Because sources are sampled uniformly during training, the expected compute
devoted to one source is
\[
x=\widehat C_{t,m}/N,
\]
where \(N\) is the number of sources in the corpus. This quantity is calculated
from the training configuration rather than measured directly, and it does
not record the exact order in which sources were sampled.

PEMS04 is scored by MAE over all valid issue times, forecast horizons and
nodes. Its 95\% intervals use 2,000 paired circular moving-block bootstrap
replicates over issue times with a 288-issue (one-day) block. CAMELS-US is
scored by median basin NSE. Its intervals use 2,000 replicates that combine
paired 30-day circular blocks over dates with independent resampling of
basins.

For each model trained on a mixed corpus, we linearly interpolate the
performance of the model trained on the evaluated source alone to estimate
what it had achieved after the same amount of compute was devoted to that
source. We compare the mixed-corpus result with both this estimate and the
final result of training on that source alone. This comparison tests how well
performance on a represented system is preserved when the system receives a
smaller share of the total training budget.

% GraphCast-style self-contained bibliography for reviewed Supplementary Material.
\phantomsection
\addcontentsline{toc}{section}{Supplementary References}
\renewcommand{\refname}{Supplementary References}
% Pre-generated from references.bib for the combined arXiv build.

\end{bibunit}

\end{document}